\def\year{2018}\relax 

\documentclass[letterpaper]{article} 
\usepackage{aaai18}  
\usepackage{times}  
\usepackage{helvet}  
\usepackage{courier}  
\usepackage{url}  
\usepackage{graphicx}  
\usepackage{amsmath}
\usepackage{bm}
\usepackage{booktabs}
\usepackage{subcaption}
\usepackage{multirow}
\usepackage{makecell}
\usepackage [english]{babel}
\newcolumntype{x}[1]{>{\centering\arraybackslash\hspace{0pt}}p{#1}}

\title{FiLM: Visual Reasoning with a General Conditioning Layer}
\author{Ethan Perez\textnormal{\textsuperscript{1,2}},
Florian Strub\textnormal{\textsuperscript{4}},
Harm de Vries\textnormal{\textsuperscript{1}},
Vincent Dumoulin\textnormal{\textsuperscript{1}},
Aaron Courville\textnormal{\textsuperscript{1,3}}\\
\textsuperscript{1}MILA, Universit\'e de Montr\'eal,
\textsuperscript{2}Rice University,
\textsuperscript{3}CIFAR Fellow,\\
\textsuperscript{4}Univ. Lille, CNRS, Centrale Lille, Inria, UMR 9189 CRIStAL France\\
ethanperez@rice.edu, florian.strub@inria.fr, mail@harmdevries.com,\{dumouliv,courvila\}@iro.umontreal.ca
}

\begin{document}
\maketitle
\begin{abstract}
    	We introduce a general-purpose conditioning method for neural networks called \textbf{FiLM}: \textbf{F}eature-w\textbf{i}se \textbf{L}inear \textbf{M}odulation. FiLM layers influence neural network computation via a simple, feature-wise affine transformation based on conditioning information. We show that FiLM layers are highly effective for visual reasoning --- answering image-related questions which require a multi-step, high-level process --- a task which has proven difficult for standard deep learning methods that do not explicitly model reasoning. Specifically, we show on visual reasoning tasks that FiLM layers 1) halve state-of-the-art error for the CLEVR benchmark, 2) modulate features in a coherent manner, 3) are robust to ablations and architectural modifications, and 4) generalize well to challenging, new data from few examples or even zero-shot.
\end{abstract}

\section{Introduction} \label{introduction}
    The ability to reason about everyday visual input is a fundamental building block of human intelligence. Some have argued that for artificial agents to learn this complex, structured process, it is necessary to build in aspects of reasoning, such as compositionality~\cite{N2NMN,IEP} or relational computation~\cite{RN}. However, if a model made from general-purpose components could learn to visually reason, such an architecture would likely be more widely applicable across domains.

\begin{figure}[t]
	\centering
    \begin{subfigure}[t]{.2\textwidth}
        \centering
        \includegraphics[height=0.8in]{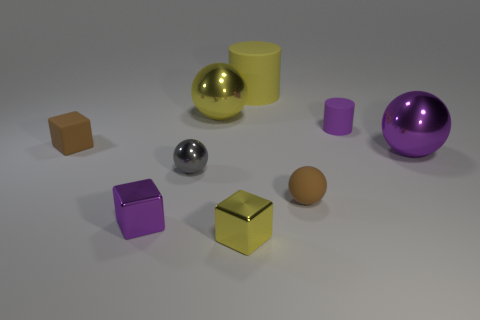}
        \caption{\bf{Q:} \it{What number of cylinders are small purple things or yellow rubber things?
        \bf{A:} \textit{2}}}
    \end{subfigure}
    ~ 
    \begin{subfigure}[t]{.2\textwidth}
		\centering
        \includegraphics[height=0.8in]{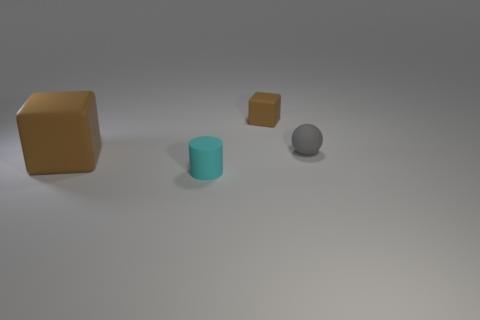}
        \caption{\bf{Q:} \it{What color is the other object that is the same shape as the large brown matte thing?
        \bf{A:} \textit{Brown}}}
    \end{subfigure}
    \caption{CLEVR examples and FiLM model answers.}
    \label{fig:CLEVR}
\end{figure}

    To understand if such a general-purpose architecture exists, we take advantage of the recently proposed CLEVR dataset~\cite{CLEVR} that tests visual reasoning via question answering. Examples from CLEVR are shown in Figure~\ref{fig:CLEVR}. Visual question answering, the general task of asking questions about images, has its own line of datasets~\cite{malinowski2014multi,geman2015visual,antol2015} which generally focus on asking a diverse set of simpler questions on images, often answerable in a single glance. From these datasets, a number of effective, general-purpose deep learning models have emerged for visual question answering~\cite{malinowski2015ask,SANs,lu2016hierarchical,anderson2017bottom}. However, tests on CLEVR show that these general deep learning approaches struggle to learn structured, multi-step reasoning~\cite{CLEVR}. In particular, these methods tend to exploit biases in the data rather than capture complex underlying structure behind reasoning~\cite{goyal2016making}.

    In this work, we show that a general model architecture can achieve strong visual reasoning with a method we introduce as \textbf{FiLM}: \textbf{F}eature-w\textbf{i}se \textbf{L}inear \textbf{M}odulation. A FiLM layer carries out a simple, feature-wise affine transformation on a neural network's intermediate features, conditioned on an arbitrary input. In the case of visual reasoning, FiLM layers enable a Recurrent Neural Network (RNN) over an input question to influence Convolutional Neural Network (CNN) computation over an image. This process adaptively and radically alters the CNN's behavior as a function of the input question, allowing the overall model to carry out a variety of reasoning tasks, ranging from counting to comparing, for example. FiLM can be thought of as a generalization of Conditional Normalization, which has proven highly successful for image stylization~\cite{CIN,CIN2,AIN}, speech recognition~\cite{DynamicLayerNorm}, and visual question answering~\cite{modulating_vision}, demonstrating FiLM's broad applicability.

    In this paper, which expands upon a shorter report~\cite{LVRWSP}, our key contribution is that we show FiLM is a strong conditioning method by showing the following on visual reasoning tasks:
    \begin{enumerate}
    	\item FiLM models achieve state-of-the-art across a variety of visual reasoning tasks, often by significant margins.
        \item FiLM operates in a coherent manner. It learns a complex, underlying structure and manipulates the conditioned network's features in a selective manner. It also enables the CNN to properly localize question-referenced objects.
        \item FiLM is robust; many FiLM model ablations still outperform prior state-of-the-art. Notably, we find there is no close link between normalization and the success of a conditioned affine transformation, a previously untouched assumption. Thus, we relax the conditions under which this method can be applied.
        \item FiLM models learn from little data to generalize to more complex and/or substantially different data than seen during training. We also introduce a novel FiLM-based zero-shot generalization method that further improves and validates FiLM's generalization capabilities.
    \end{enumerate}

\section{Method} \label{method}

\begin{figure}[t]
            \centering
            \includegraphics[width=0.45\linewidth]{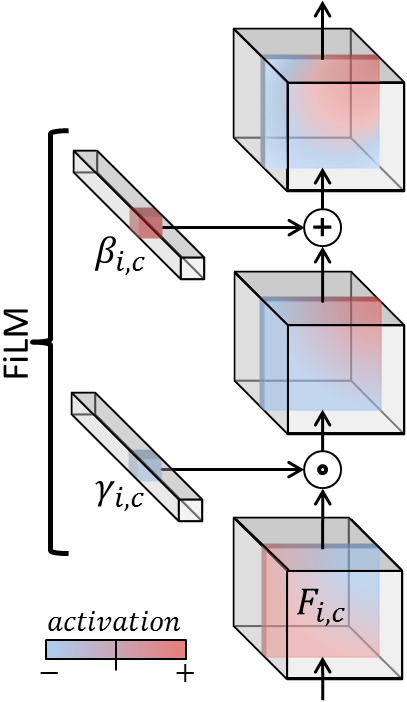}
	            \caption{A single FiLM layer for a CNN. The dot signifies a Hadamard product. Various combinations of $\gamma$ and $\beta$ can modulate individual feature maps in a variety of ways.}
            \label{fig:FiLM}
        \end{figure}

    Our model processes the question-image input using FiLM, illustrated in Figure~\ref{fig:FiLM}. We start by explaining FiLM and then describe our particular model for visual reasoning.

	\subsection{Feature-wise Linear Modulation}
		\label{FiLM}
FiLM learns to adaptively influence the output of a neural network by applying an affine transformation, or FiLM, to the network's intermediate features, based on some input. More formally, FiLM learns
functions $f$ and $h$ which output $\gamma_{i,c}$ and $\beta_{i,c}$ as a function of input $\bm{x_i}$:
  \begin{align}
  	\gamma_{i,c} = f_c(\bm{x}_i) \qquad ~ \qquad
  	\beta_{i,c} = h_c(\bm{x}_i),
  \end{align}
where $\gamma_{i,c}$ and $\beta_{i,c}$ modulate a neural network's activations $\bm{F}_{i,c}$, whose subscripts refer to the $i^{th}$ input's $c^{th}$ feature or feature map, via a feature-wise affine transformation:
  \begin{equation}
  	FiLM(\bm{F}_{i,c} | \gamma_{i,c}, \beta_{i,c}) = \gamma_{i,c}\bm{F}_{i,c} + \beta_{i,c}.
  \end{equation}
$f$ and $h$ can be arbitrary functions such as neural networks. Modulation of a target neural network's processing can be based on the same input to that neural network or some other input, as in the case of multi-modal or conditional tasks. For CNNs, $f$ and $h$ thus modulate the per-feature-map distribution of activations based on $\bm{x_i}$, agnostic to spatial location.

	In practice, it is easier to refer to $f$ and $h$ as a single function that outputs one $(\bm{\gamma}, \bm{\beta})$ vector, since, for example, it is often beneficial to share parameters across $f$ and $h$ for more efficient learning. We refer to this single function as the FiLM generator. We also refer to the network to which FiLM layers are applied as the Feature-wise Linearly Modulated network, the FiLM-ed network.

	FiLM layers empower the FiLM generator to manipulate feature maps of a target, FiLM-ed network by scaling them up or down, negating them, shutting them off, selectively thresholding them (when followed by a ReLU), and more. Each feature map is conditioned independently, giving the FiLM generator moderately fine-grained control over activations at each FiLM layer.
    
    As FiLM only requires two parameters per modulated feature map, it is a scalable and computationally efficient conditioning method. In particular, FiLM has a computational cost that does not scale with the image resolution.

	\subsection{Model}
		\label{model}

        \begin{figure}[t]
            \centering
            \includegraphics[width=0.9\linewidth]{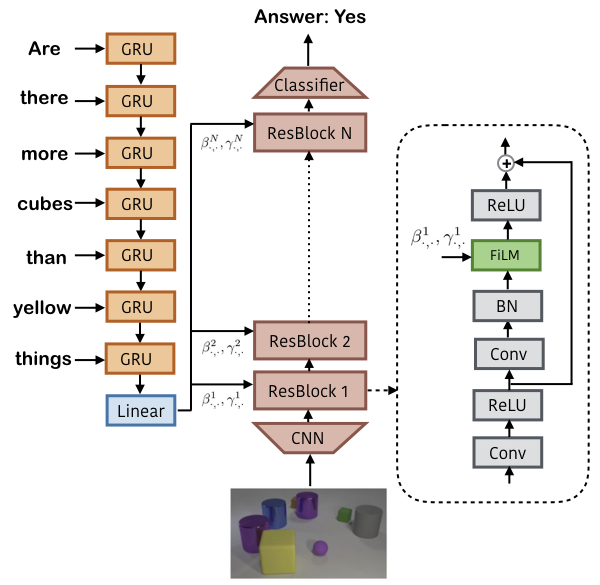}
	            \caption{The FiLM generator (left), FiLM-ed network (middle), and residual block architecture (right) of our model.}
            \label{fig:model}
        \end{figure}

		Our FiLM model consists of a FiLM-generating linguistic pipeline and a FiLM-ed visual pipeline as depicted in Figure~\ref{fig:model}. The FiLM generator processes a question $\bm{x_i}$ using a Gated Recurrent Unit (GRU) network~\cite{GRU} with 4096 hidden units that takes in learned, 200-dimensional word embeddings. The final GRU hidden state is a question embedding, from which the model predicts $(\bm{\gamma}^{n}_{i,\cdot}, \bm{\beta}^{n}_{i,\cdot})$ for each $n^{th}$ residual block via affine projection.

        The visual pipeline extracts 128 $14\times14$ image feature maps from a resized, $224\times224$ image input using either a CNN trained from scratch or a fixed, pre-trained feature extractor with a learned layer of $3\times3$ convolutions. The CNN trained from scratch consists of 4 layers with 128 $4\times4$ kernels each, ReLU activations, and batch normalization, similar to prior work on CLEVR~\cite{RN}. The fixed feature extractor outputs the \textit{conv4} layer of a ResNet-101~\cite{ResNet} pre-trained on ImageNet \cite{ImageNet} to match prior work on CLEVR~\cite{CLEVR,IEP}. Image features are processed by several --- 4 for our model --- FiLM-ed residual blocks (ResBlocks) with 128 feature maps and a final classifier. The classifier consists of a $1\times1$ convolution to 512 feature maps, global max-pooling, and a two-layer MLP with 1024 hidden units that outputs a softmax distribution over final answers.

        Each FiLM-ed ResBlock starts with a $1\times1$ convolution followed by one $3\times3$ convolution with an architecture as depicted in Figure~\ref{fig:model}. We turn the parameters of batch normalization layers that immediately precede FiLM layers off. Drawing from prior work on CLEVR~\cite{N2NMN,RN} and visual reasoning~\cite{VIN}, we concatenate two coordinate feature maps indicating relative x and y spatial position (scaled from $-1$ to $1$) with the image features, each ResBlock's input, and the classifier's input to facilitate spatial reasoning.
        
        We train our model end-to-end from scratch with Adam~\cite{Adam} (learning rate $3e^{-4}$), weight decay ($1e^{-5}$), batch size 64, and batch normalization and ReLU throughout FiLM-ed network. Our model uses only image-question-answer triplets from the training set without data augmentation. We employ early stopping based on validation accuracy, training for 80 epochs maximum. Further model details are in the appendix. Empirically, we found FiLM had a large capacity, so many architectural and hyperparameter choices were for added regularization.
        
        We stress that our model relies \textit{solely} on feature-wise affine conditioning to use question information influence the visual pipeline behavior to answer questions. This approach differs from classical visual question answering pipelines which fuse image and language information into a single embedding via element-wise product, concatenation, attention, and/or more advanced methods~\cite{SANs,lu2016hierarchical,anderson2017bottom}.
        
\section{Related Work}\label{relatedwork}

	FiLM can be viewed as a generalization of Conditional Normalization (CN) methods. CN replaces the parameters of the feature-wise affine transformation typical in normalization layers, as introduced originally~\cite{BN}, with a learned function of some conditioning information. Various forms of CN have proven highly effective across a number of domains: Conditional Instance Norm~\cite{CIN,CIN2} and Adaptive Instance Norm~\cite{AIN} for image stylization, Dynamic Layer Norm for speech recognition~\cite{DynamicLayerNorm}, and Conditional Batch Norm for general visual question answering on complex scenes such as VQA and GuessWhat?!~\cite{modulating_vision}. This work complements our own, as we seek to show that feature-wise affine conditioning is effective for multi-step reasoning and understand the underlying mechanism behind its success.
    
    Notably, prior work in CN has not examined whether the affine transformation must be placed directly after normalization. Rather, prior work includes normalization in the method name for instructive purposes or due to implementation details. We investigate the connection between FiLM and normalization, finding it not strictly necessary for the affine transformation to occur directly after normalization. Thus, we provide a unified framework for all of these methods through FiLM, as well as a normalization-free relaxation of this approach which can be more broadly applied.

	Beyond CN, there are many connections between FiLM and other conditioning methods. A common approach, used for example in Conditional DCGANs~\cite{DCGAN}, is to concatenate constant feature maps of conditioning information with convolutional layer input. Though not as parameter efficient, this method simply results in a feature-wise conditional bias. Likewise, concatenating conditioning information with fully-connected layer input amounts to a feature-wise conditional bias. Other approaches such as WaveNet~\cite{WaveNet} and Conditional PixelCNN~\cite{van2016conditional} directly add a conditional feature-wise bias. These approaches are equivalent to FiLM with $\bm{\gamma}=\bm{1}$, which we compare FiLM to in the Experiments section. In reinforcement learning, an alternate formulation of FiLM has been used to train one game-conditioned deep Q-network to play ten Atari games~\cite{OCF}, though FiLM was neither the focus of this work nor analyzed as a major component.
    
    Other methods gate an input's features as a function of that same input, rather than a separate conditioning input. These methods include LSTMs for sequence modeling~\cite{LSTM}, Convolutional Sequence to Sequence for machine translation~\cite{pmlr-v70-gehring17a}, and even the ImageNet 2017 winning model, Squeeze and Excitation Networks~\cite{hu2017}. This approach amounts to a feature-wise, conditional scaling, restricted to between 0 and 1, while FiLM consists of both scaling and shifting, each unrestricted. In the Experiments section, we show the effect of restricting FiLM's scaling to between 0 and 1 for visual reasoning. We find it noteworthy that this general approach of feature modulation is effective across a variety of settings and architectures.
    
        \begin{table*}[t]
        \centering
        
        {\small
        \begin{tabular}{l|c|cccccc}
        \toprule
        {Model} & {\textbf{Overall}} & {Count} & {Exist} & \begin{tabular}{@{}c@{}}Compare \\ Numbers\end{tabular} & \begin{tabular}{@{}c@{}}Query \\ Attribute\end{tabular} & \begin{tabular}{@{}c@{}}Compare \\ Attribute\end{tabular}\\
        \midrule
        Human~\cite{IEP}                   & 92.6 &86.7 &96.6 &86.5 &95.0 &96.0\\
        \midrule
        Q-type baseline~\cite{IEP}         &41.8 &34.6 &50.2 &51.0 &36.0 &51.3\\
        LSTM~\cite{IEP}                    &46.8 &41.7 &61.1 &69.8 &36.8 &51.8\\
        CNN+LSTM~\cite{IEP}                &52.3 &43.7 &65.2 &67.1 &49.3 &53.0\\
        CNN+LSTM+SA~\cite{RN}              &76.6 &64.4 &82.7 &77.4 &82.6 &75.4\\
        N2NMN*~\cite{N2NMN}                &83.7 &68.5 &85.7 &84.9 &90.0 &88.7\\
        PG+EE (9K prog.)*~\cite{IEP}       &88.6 &79.7 &89.7 &79.1 &92.6 &96.0\\
        PG+EE (700K prog.)*~\cite{IEP}     &96.9 &92.7 &97.1 &\bf{98.7} &98.1 &98.9\\
        CNN+LSTM+RN\textdagger$\ddagger$~\cite{RN}  &95.5 &90.1 &97.8 & 93.6 &97.9 &97.1\\
        \midrule
        CNN+GRU+FiLM &\bf{97.7} &\bf{94.3} &99.1 &96.8 &99.1 &99.1\\
        CNN+GRU+FiLM$\ddagger$ &97.6 &\bf{94.3} &\bf{99.3} &93.4 &\bf{99.3} &\bf{99.3}\\
         \bottomrule
        \end{tabular}
        }
        \caption{\label{tab:results}
        CLEVR accuracy (overall and per-question-type) by baselines, competing methods, and FiLM. (*) denotes use of extra supervision via program labels. (\textdagger) denotes use of data augmentation. ($\ddagger$) denotes training from raw pixels.}
        \end{table*}
    
        There are even broader links between FiLM and other methods. For example, FiLM can be viewed as using one network to generate parameters of another network, making it a form of hypernetwork~\cite{Hypernets}. Also, FiLM has potential ties with conditional computation and mixture of experts methods, where specialized network sub-parts are active on a per-example basis~\cite{HME,DeepMoE,shazeer2017outrageously}; we later provide evidence that FiLM learns to selectively highlight or suppress feature maps based on conditioning information. Those methods select at a sub-network level while FiLM selects at a feature map level.

	In the domain of visual reasoning, one leading method is the Program Generator + Execution Engine model~\cite{IEP}. This approach consists of a sequence-to-sequence Program Generator, which takes in a question and outputs a sequence corresponding to a tree of composable neural modules, each of which is a two or three layer residual block. This tree of neural modules is assembled to form the Execution Engine that then predicts an answer from the image. This modular approach is part of a line of neural module network methods~\cite{NMNQA,NMN,N2NMN}, of which End-to-End Module Networks~\cite{N2NMN} have also been tested on visual reasoning. These models use strong priors by explicitly modeling the compositional nature of reasoning and by training with additional program labels, \textit{i.e.} ground-truth step-by-step instructions on how to correctly answer a question. End-to-End Module Networks further build in model biases via per-module, hand-crafted neural architectures for specific functions. Our approach learns directly from visual and textual input without additional cues or a specialized architecture.

	Relation Networks (RNs) are another leading approach for visual reasoning~\cite{RN}. RNs succeed by explicitly building in a comparison-based prior. RNs use an MLP to carry out pairwise comparisons over each location of extracted convolutional features over an image, including LSTM-extracted question features as input to this MLP. RNs then element-wise sum over the resulting comparison vectors to form another vector from which a final classifier predicts the answer. We note that RNs have a computational cost that scales quadratically in spatial resolution, while FiLM's cost is independent of spatial resolution. Notably, since RNs concatenate question features with MLP input, a form of feature-wise conditional biasing as explained earlier, their conditioning approach is related to FiLM.

\section{Experiments} \label{experiments}

    First, we test our model on visual reasoning with the CLEVR task and use trained FiLM models to analyze what FiLM learns. Second, we explore how well our model generalizes to more challenging questions with the CLEVR-Humans task. Finally, we examine how FiLM performs in few-shot and zero-shot generalization settings using the CLEVR Compositional Generalization Test. In the appendix, we provide an error analysis of our model. Our code is available at~\url{https://github.com/ethanjperez/film}.

	\subsection{CLEVR Task} \label{CLEVR}
    	CLEVR is a synthetic dataset of 700K (image, question, answer, program) tuples~\cite{CLEVR}. Images contain 3D-rendered objects of various shapes, materials, colors, and sizes. Questions are multi-step and compositional in nature, as shown in Figure~\ref{fig:CLEVR}. They range from counting questions (\textit{``How many green objects have the same size as the green metallic block?''}) to comparison questions (\textit{``Are there fewer tiny yellow cylinders than yellow metal cubes?''}) and can be 40+ words long. Answers are each one word from a set of $28$ possible answers. Programs are an additional supervisory signal consisting of step-by-step instructions, such as \texttt{filter\_shape[cube]}, \texttt{relate[right]}, and \texttt{count}, on how to answer the question.
        
\begin{figure*}[ht!]
   		\small{
       \begin{tabular}{m{3.1cm}m{3.1cm}m{3.1cm}m{3.1cm}m{3.1cm}}
  	  \textbf{Q:} \textit{What shape is the...}
  	  &
      \textit{...purple thing?} 
      \textbf{A:} \textit{cube}
       &
       \textit{...blue thing?} 
       \textbf{A:} \textit{sphere}
       &
       \textit{...red thing right of the blue thing?} 
       \textbf{A:} \textit{sphere}
       &
       \textit{...red thing left of the blue thing?} 
       \textbf{A:} \textit{cube}
       \\
      \includegraphics[width=\linewidth]{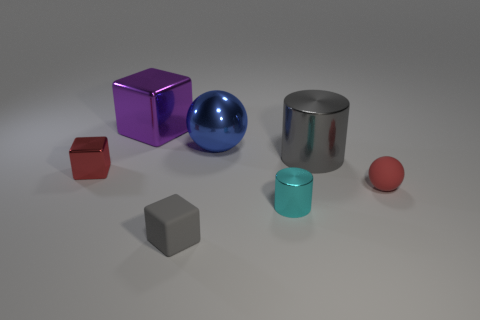}
	  & 	
  	  \includegraphics[width=\linewidth]{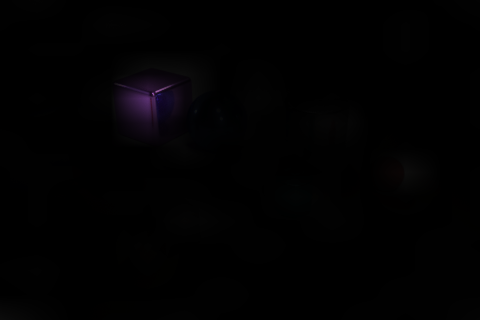}
	  &
  	  \includegraphics[width=\linewidth]{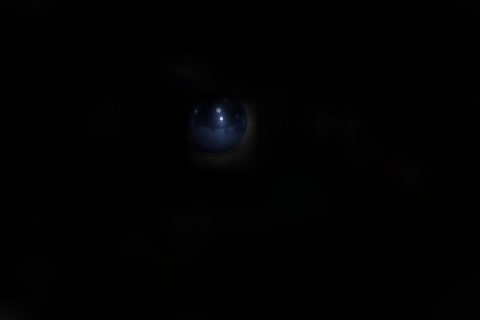}
	  &
\includegraphics[width=\linewidth]{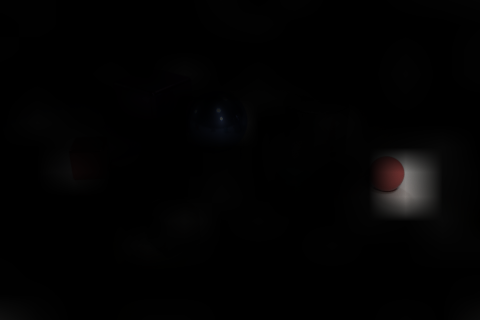}
	  &
  	  \includegraphics[width=\linewidth]{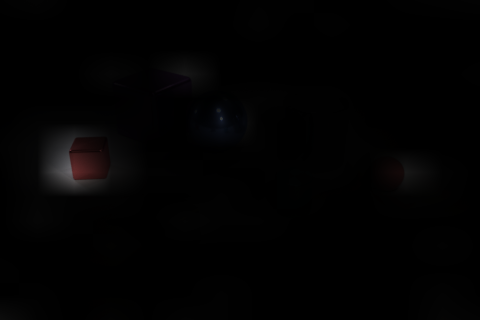}
    \\
      	  \includegraphics[width=\linewidth]{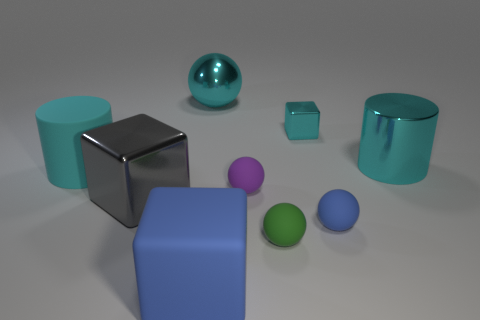}
&
  	  \includegraphics[width=\linewidth]{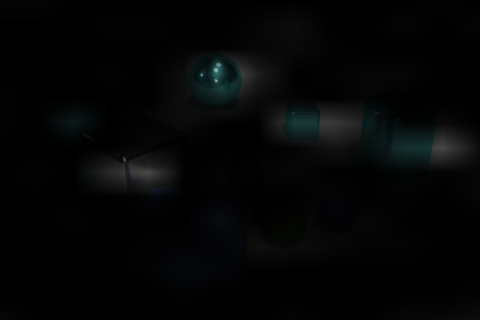}
&
  	  \includegraphics[width=\linewidth]{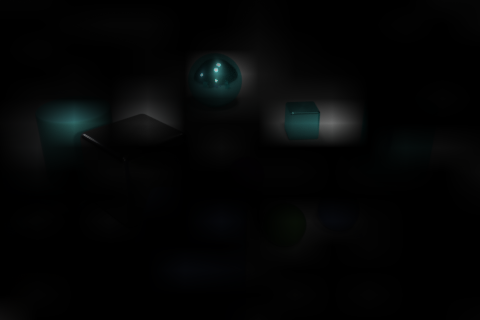}
&
  	  \includegraphics[width=\linewidth]{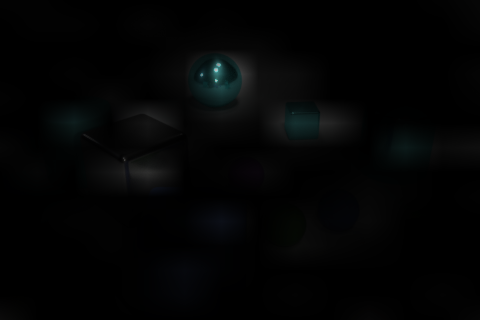}
&
  	  \includegraphics[width=\linewidth]{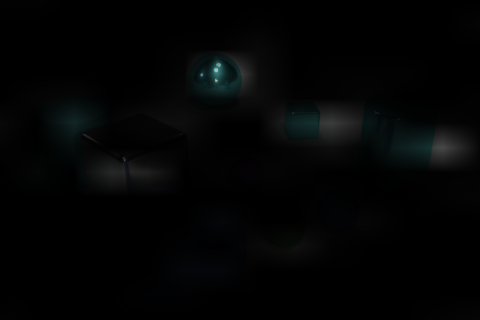}
\cr
	\\
    \textbf{Q:} \textit{How many cyan things are...}
    &
    \textit{...right of the gray cube?}
    \textbf{A:} \textit{3}
    &
    \textit{...left of the small cube?}
    \textbf{A:} \textit{2}
    &
    \textit{...right of the gray cube and left of the small cube?}
    \textbf{A:} \textit{1}
    &
    \textit{...right of the gray cube or left of the small cube? }     
    \textbf{A:} \textit{4} (\textbf{P:} \textit{3})
	\end{tabular}
    }
    \caption{Visualizations of the distribution of locations which the model uses for its globally max-pooled features which its final MLP predicts from. FiLM correctly localizes the answer-referenced object (top) or all question-referenced objects (bottom), but not as accurately when it answers incorrectly (rightmost bottom). Questions and images used match~\cite{IEP}.}
      	\label{fig:pool-feature-locations}
  \end{figure*}
        
    \subsubsection{Baselines} \label{baselines}
    	We compare against the following methods, discussed in detail in the Related Work section:
        \begin{itemize}
        	\item \textbf{Q-type baseline}: Predicts based on a question's category.
            \item \textbf{LSTM}: Predicts using only the question. 
            \item \textbf{CNN+LSTM}: MLP prediction over CNN-extracted image features and LSTM-extracted question features.
        	\item \textbf{Stacked Attention Networks (CNN+LSTM+SA)}: Linear prediction over CNN-extracted image feature and LSTM-extracted question features combined via two rounds of soft spatial attention~\cite{SANs}.
             \item \textbf{End-to-End Module Networks (N2NMN) and Program Generator + Execution Engine (PG+EE)}: Methods in which separate neural networks learn separate sub-functions and are assembled into a question-dependent structure~\cite{N2NMN,IEP}.
             \item \textbf{Relation Networks (CNN+LSTM+RN)}: An approach which builds in pairwise comparisons over spatial locations to explicitly model reasoning's relational nature~\cite{RN}.
        \end{itemize}

	\subsubsection{Results} \label{results}

		FiLM achieves a new overall state-of-the-art on CLEVR, as shown in Table~\ref{tab:results}, outperforming humans and previous methods, including those using explicit models of reasoning, program supervision, and/or data augmentation. For methods not using extra supervision, FiLM roughly halves state-of-the-art error (from $4.5\%$ to $2.3\%$). Note that using pre-trained image features as input can be viewed as a form of data augmentation in itself but that FiLM performs equally well using raw pixel inputs. Interestingly, the raw pixel model seems to perform better on lower-level questions (\textit{i.e.} querying and comparing attributes) while the image features model seems to perform better on higher-level questions (\textit{i.e.} compare numbers of objects).

	\subsection{What Do FiLM Layers Learn?} \label{what-is-learned}
        
        To understand how FiLM visually reasons, we visualize activations to observe the net result of FiLM layers. We also use histograms and t-SNE~\cite{maaten2008visualizing} to find patterns in the learned FiLM $\gamma$ and $\beta$ parameters themselves. In Figures~\ref{fig:film-effect-1} and~\ref{fig:film-effect-2} in the appendix, we visualize the effect of FiLM at the single feature map level.
        
        \subsubsection{Activation Visualizations}
        Figure~\ref{fig:pool-feature-locations} visualizes the distribution of locations responsible for the globally-pooled features which the MLP in the model's final classifier uses to predict answers. These images reveal that the FiLM model predicts using features of areas near answer-related or question-related objects, as the high CLEVR accuracy also suggests. This finding highlights that appropriate feature modulation indirectly results in spatial modulation, as regions with question-relevant features will have large activations while other regions will not. This observation might explain why FiLM outperforms Stacked Attention, the next best method not explicitly built for reasoning, so significantly ($21\%$); FiLM appears to carry many of spatial attention's benefits, while also influencing feature representation.
        
		Figure~\ref{fig:pool-feature-locations} also suggests that the FiLM-ed network carries out reasoning throughout its pipeline. In the top example, the FiLM-ed network has localized the answer-referenced object alone before the MLP classifier. In the bottom example, the FiLM-ed network retains, for the MLP classifier, features on objects that are not referred to by the answer but are referred to by the question. The latter example provides evidence that the final MLP itself carries out some reasoning, using FiLM to extract relevant features for its reasoning.
        
        \begin{figure}
          \begin{subfigure}[b]{0.23\textwidth}
              \includegraphics[width=\linewidth]{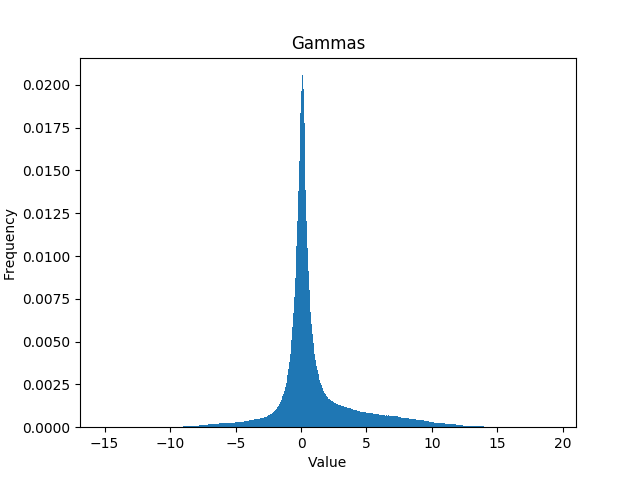}
          \end{subfigure}%
          \begin{subfigure}[b]{0.23\textwidth}
              \includegraphics[width=\linewidth]{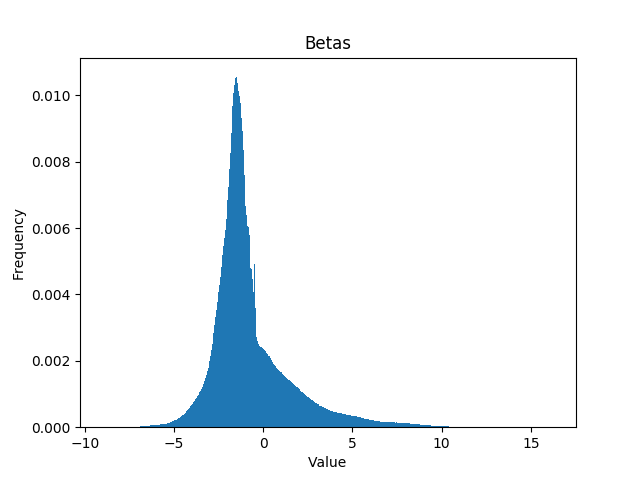}
          \end{subfigure}%
        \caption{Histograms of $\gamma_{i,c}$ (left) and $\beta_{i,c}$ (right) values over all FiLM layers, calculated over the validation set.}
        \label{fig:gammas-betas}
        \end{figure}
        
        \begin{figure*}[t]
        \centering
        \includegraphics[width=0.9\linewidth]{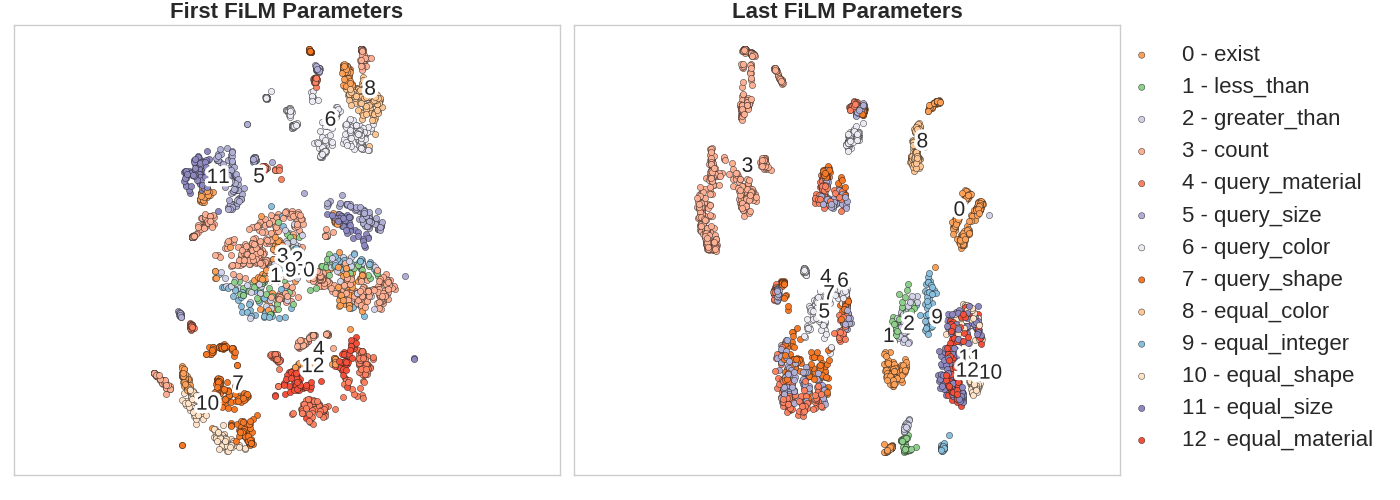}
        \caption{t-SNE plots of $(\bm{\gamma}$, $\bm{\beta})$ of the first (left) and last (right) FiLM layers of a 6-FiLM layer Network. FiLM parameters cluster by low-level reasoning functions in the first layer and by high-level reasoning functions in the last layer.}
        \label{fig:tsne}
        \end{figure*}
        
        \subsubsection{FiLM Parameter Histograms}
		To analyze at a lower level how FiLM uses the question to condition the visual pipeline, we plot $\gamma$ and $\beta$ values predicted over the validation set, as shown in Figure~\ref{fig:gammas-betas} and in more detail in the appendix (Figures~\ref{fig:gammas-layerwise} to~\ref{fig:film-stats}). $\gamma$ and $\beta$ values take advantage of a sizable range, varying from -15 to 19 and from -9 to 16, respectively. $\gamma$ values show a sharp peak at 0, showing that FiLM learns to use the question to shut off or significantly suppress whole feature maps. Simultaneously, FiLM learns to upregulate a much more selective set of other feature maps with high magnitude $\gamma$ values. Furthermore, a large fraction ($36\%$) of $\gamma$ values are negative; since our model uses a ReLU after FiLM, $\gamma<0$ can cause a significantly different set of activations to pass the ReLU to downstream layers than $\gamma>0$. Also, $76\%$ of $\beta$ values are negative, suggesting that FiLM also uses $\beta$ to be selective about which activations pass the ReLU. We show later that FiLM's success is largely architecture-agnostic, but examining a particular model gives insight into the influence FiLM learns to exert in a specific case. Together, these findings suggest that FiLM learns to selectively upregulate, downregulate, and shut off feature maps based on conditioning information.
        
        \subsubsection{FiLM Parameters t-SNE Plot}
        In Figure~\ref{fig:tsne}, we visualize FiLM parameter vectors $(\bm{\gamma}, \bm{\beta})$ for 3,000 random validation points with t-SNE. We analyze the deeper, 6-ResBlock version of our model, which has a similar validation accuracy as our 4-ResBlock model, to better examine how FiLM layers in different layers of a hierarchy behave. First and last layer FiLM $(\bm{\gamma}, \bm{\beta})$ are grouped by the low-level and high-level reasoning functions necessary to answer CLEVR questions, respectively. For example, FiLM parameters for \texttt{equal\_color} and \texttt{query\_color} are close for the first layer but apart for the last layer. The same is true for shape, size and material questions. Conversely, \texttt{equal\_shape}, \texttt{equal\_size}, and \texttt{equal\_material} FiLM parameters are grouped in the last layer but split in the first layer --- likewise for other high level groupings such as integer comparison and querying.
        These findings suggest that FiLM layers learn a sort of function-based modularity without an architectural prior. Simply with end-to-end training, FiLM learns to handle not only different types of questions differently, but also different types of question sub-parts differently; the FiLM model works from low-level to high-level processes as is the proper approach. For models with fewer FiLM layers, such patterns also appear, but less clearly; these models must begin higher level reasoning sooner.

	\subsection{Ablations} \label{ablations}
    
\begin{figure}[ht]
        \centering
        \includegraphics[width=0.8\linewidth]{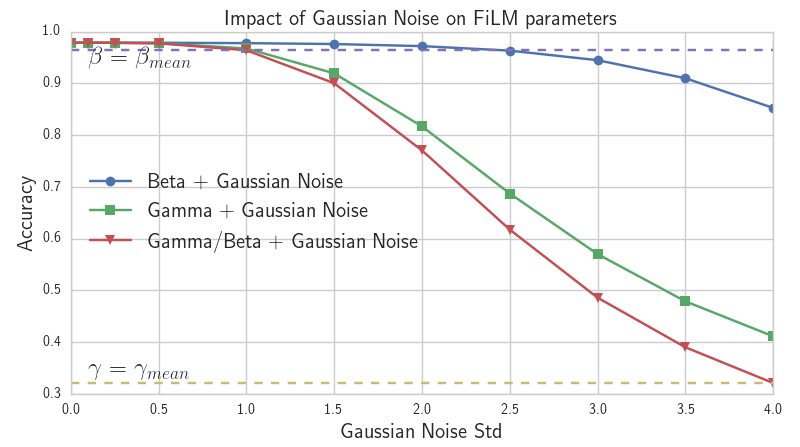}
        \caption{\label{fig:noisy-ablations} An analysis of how robust FiLM parameters are to noise at test time. The horizontal lines correspond to setting $\gamma$ or $\beta$ to their respective training set mean values.}
        \end{figure}

    	Using the validation set, we conduct an ablation study on our best model to understand how FiLM learns visual reasoning. We show results for test time ablations in Figure~\ref{fig:noisy-ablations}, for architectural ablations in Table~\ref{tab:ablations}, and for varied model depths in Table~\ref{tab:num-resblocks}. Without hyperparameter tuning, most architectural ablations and model depths outperform prior state-of-the-art on training from only image-question-answer triplets, supporting FiLM's overall robustness. Table~\ref{tab:num-resblocks} also shows using the validation set that our results are statistically significant.
    
\begin{table}[ht!]
        {\small
       \begin{tabular}{lm{4.9cm}|x{1.4cm}}
        \toprule
        {Model}   && {\textbf{Overall}} \\
        \end{tabular}
        \\	
        \begin{tabular}{lm{5.7cm}|x{1.4cm}}
        \midrule
        \multicolumn{2}{l|}{\textbf{Restricted $\bm{\gamma}$ or $\bm{\beta}$}}&\\
        & FiLM with $\bm{\beta}:=\bm{0}$ &96.9\\
        & FiLM with $\bm{\gamma}:=\bm{1}$ &95.9\\
        & FiLM with $\bm{\gamma}:=\sigma(\bm{\gamma})$ &95.9\\
        & FiLM with $\bm{\gamma}:=tanh(\bm{\gamma})$ &96.3\\
        & FiLM with $\bm{\gamma}:=exp(\bm{\gamma})$ &96.3\\
        \end{tabular}
        \\	
        \begin{tabular}{lm{5.7cm}|x{1.4cm}}
        \midrule
        \multicolumn{2}{l|}{\textbf{Moving FiLM within ResBlock}}&\\
        &FiLM after residual connection &96.6\\
        &FiLM after ResBlock ReLU-2 &97.7\\
        &FiLM after ResBlock Conv-2 &97.1\\
        &FiLM before ResBlock Conv-1 &95.0\\
        \end{tabular}
        \\	
        \begin{tabular}{lm{5.7cm}|x{1.4cm}}
        \midrule
        \multicolumn{2}{l|}{\textbf{Removing FiLM from ResBlocks}}&\\
        &No FiLM in ResBlock 4 &96.8\\
        &No FiLM in ResBlock 3-4 &96.5\\
        &No FiLM in ResBlock 2-4 &97.3\\
        &No FiLM in ResBlock 1-4 &21.4\\
        \end{tabular}
        \\	
        \begin{tabular}{lm{5.7cm}|x{1.4cm}}
        \midrule
        \multicolumn{2}{l|}{\textbf{Miscellaneous}}&\\
        &$1\times1$ conv only, with no coord. maps &95.3\\
        &No residual connection &94.0\\
        &No batch normalization &93.7\\
        &Replace image features with raw pixels & 97.6\\
        \end{tabular}
        \\
        \begin{tabular}{lm{5.7cm}|x{1.4cm}}
        \midrule
        & Best Architecture &97.4$\pm$.4\\ 
        \bottomrule
        
        \end{tabular}
        }
        \caption{\label{tab:ablations}
        CLEVR val accuracy for ablations, trained with the best architecture with only specified changes. We report the standard deviation of the best model accuracy over 5 runs.}
        \end{table}
        
        \subsubsection{Effect of $\gamma$ and $\beta$}
        
        To test the effect of $\bm{\gamma}$ and $\bm{\beta}$ separately, we trained one model with a constant $\bm{\gamma}=\bm{1}$ and another with $\bm{\beta}=\bm{0}$. With these models, we find a 1.5\% and .5\% accuracy drop, respectively; FiLM can learn to condition the CNN for visual reasoning through either biasing or scaling alone, albeit not as well as conditioning both together. This result also suggests that $\bm{\gamma}$ is more important than $\bm{\beta}$.
        
        To further compare the importance of $\bm{\gamma}$ and $\bm{\beta}$, we run a series of test time ablations (Figure~\ref{fig:noisy-ablations}) on our best, fully-trained model. First, we replace $\bm{\beta}$ with the mean $\bm{\beta}$ across the training set. This ablation in effect removes all conditioning information from $\bm{\beta}$ parameters during test time, from a model trained to use both $\bm{\gamma}$ and $\bm{\beta}$. Here, we find that accuracy only drops by 1.0\%, while the same procedure on $\bm{\gamma}$ results in a 65.4\% drop. This large difference suggests that, in practice, FiLM largely conditions through $\bm{\gamma}$ rather than $\bm{\beta}$. Next, we analyze performance as we add increasingly more Gaussian noise to the best model's FiLM parameters at test time. Noise in gamma hurts performance significantly more, showing FiLM's higher sensitivity to changes in $\bm{\gamma}$ than in $\bm{\beta}$ and corroborating the relatively greater importance of $\bm{\gamma}$.
        
        \subsubsection{Restricting $\gamma$}
        To understand what aspect of $\bm{\gamma}$ is most effective, we train a model that limits $\bm{\gamma}$ to $(0,1)$ using sigmoid, as many models which use feature-wise, multiplicative gating do. Likewise, we also limit $\bm{\gamma}$ to $(-1,1)$ using $tanh$. Both restrictions hurt performance, roughly as much as removing conditioning from $\bm{\gamma}$ entirely by training with $\bm{\gamma}=\bm{1}$. Thus, FiLM's ability to scale features by large magnitudes appears to contribute to its success. Limiting $\bm{\gamma}$ to $(0,\infty)$ with $exp$ also hurts performance, validating the value of FiLM's capacity to negate and zero out feature maps.
        
        \subsubsection{Conditional Normalization}
        We perform an ablation study on the placement of FiLM to evaluate the relationship between normalization and FiLM that Conditional Normalization approaches assume. Unfortunately, it is difficult to accurately decouple the effect of FiLM from normalization by simply training our corresponding model without normalization, as normalization significantly accelerates, regularizes, and improves neural network learning~\cite{BN}, but we include these results for completeness. However, we find no substantial performance drop when moving FiLM layers to different parts of our model's ResBlocks; we even reach the upper end of the best model's performance range when placing FiLM after the post-normalization ReLU in the ResBlocks. Thus, we decouple the name from normalization for clarity regarding where the fundamental effectiveness of the method comes from. By demonstrating this conditioning mechanism is not closely connected to normalization, we open the doors to applications other settings in which normalization is less common, such as RNNs and reinforcement learning, which are promising directions for future work with FiLM.
        
        \subsubsection{Repetitive Conditioning}
        To understand the contribution of repetitive conditioning towards FiLM model success, we train FiLM models with successively fewer FiLM layers. Models with fewer FiLM layers, even a single FiLM layer, do not deviate far from the best model's performance, revealing that the model can reason and answer diverse questions successfully by modulating features even just once. This observation highlights the capacity of even one FiLM layer. Perhaps one FiLM layer can pass enough question information to the CNN to enable it to carry out reasoning later in the network, in place of the more hierarchical conditioning deeper FiLM models appear to use. We leave more in-depth investigation of this matter for future work.
        
        \subsubsection{Spatial Reasoning}
        To examine how FiLM models approach spatial reasoning, we train a version of our best model architecture, from image features, with only $1\times1$ convolutions and without feeding coordinate feature maps indicating relative spatial position to the model. Due to the global max-pooling near the end of the model, this model cannot transfer information across spatial positions. Notably, this model still achieves a high 95.3\% accuracy, indicating that FiLM models are able to reason about space simply from the spatial information contained in a single location of fixed image features.
        
        \subsubsection{Residual Connection}
		Removing the residual connection causes one of the larger accuracy drops. Since there is a global max-pooling operation near the end of the network, this finding suggests that the best model learns to primarily use features of locations that are repeatedly important throughout lower and higher levels of reasoning to make its final decision. The higher accuracies for models with FiLM modulating features inside residual connections rather than outside residual connections supports this hypothesis.
        
\begin{table}[t]
        \centering
        \begin{tabular}{l|c||l|cc|}
        \toprule
        {Model} & {\textbf{Overall}} & {Model} & {\textbf{Overall}} \\
        \midrule
         1 ResBlock &93.5 & 6 ResBlocks &97.7\\
         2 ResBlocks &97.1 &7 ResBlocks &97.4\\
         3 ResBlocks &96.7 &8 ResBlocks &97.6\\
         4 ResBlocks &97.4$\pm$.4 &12 ResBlocks &96.9\\
         5 ResBlocks &97.4  &  & \\
\end{tabular}
\caption{\label{tab:num-resblocks} CLEVR val accuracy by FiLM model depth.}
\end{table}

		\subsubsection{Model Depth}
        Table~\ref{tab:num-resblocks} shows model performance by the number of ResBlocks. FiLM is robust to varying depth but less so with only 1 ResBlock, backing the earlier theory that the FiLM-ed network reasons throughout its pipeline.

\begin{figure*}[t!]
		\centering
   		\small{
\begin{tabular}{m{3.1cm}m{3.1cm}m{3.1cm}m{3.1cm}m{3.1cm}}
    \\
      	  \includegraphics[width=\linewidth]{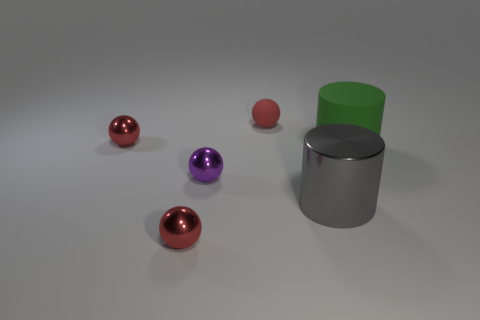}
&
  	  \includegraphics[width=\linewidth]{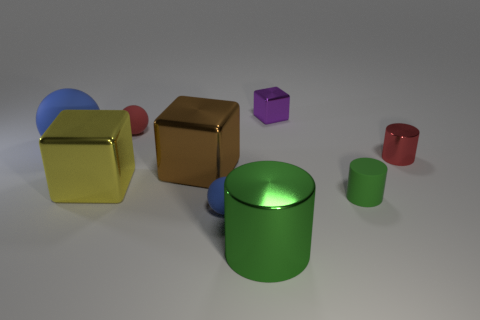}
&
  	  \includegraphics[width=\linewidth]{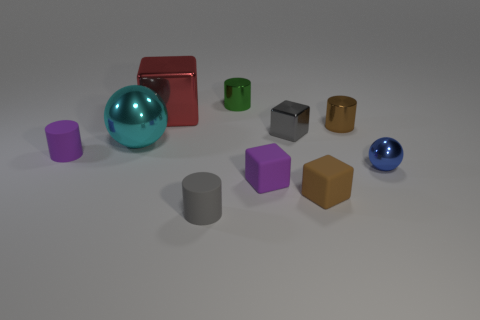}
&
  	  \includegraphics[width=\linewidth]{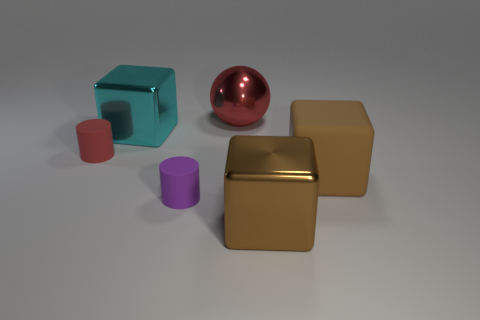}
&
  	  \includegraphics[width=\linewidth]{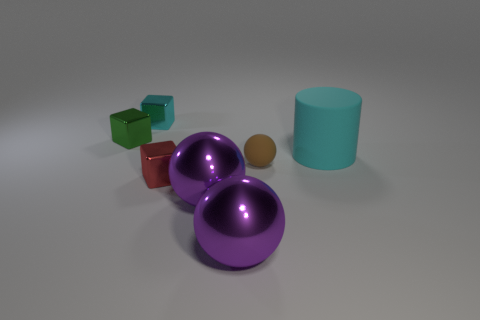}
\cr
	\\
    \textbf{Q:} \textit{What object is the color of \underline{grass}?}
    \textbf{A:} \textit{Cylinder}
    &
    \textbf{Q:} \textit{\underline{Which} shape objects are \underline{partially} \underline{obscured} \underline{from} \underline{view}?}
    \textbf{A:} \textit{Sphere}
    &
    \textbf{Q:} \textit{What color is the matte object \underline{farthest} to the right?}
    \textbf{A:} \textit{Brown}
    &
    \textbf{Q:} \textit{What shape is \underline{reflecting} in the large cube?}
    \textbf{A:} \textit{Cylinder}
    &
    \textbf{Q:} \textit{If \underline{all} \underline{cubical} objects were \underline{removed} what \underline{shaped} objects \underline{would} there \underline{be} the \underline{most} of?}     
    \textbf{A:} \textit{Sphere} (\textbf{P:} \textit{Rubber})
	\end{tabular}
    }
    \caption{Examples from CLEVR-Humans, which introduces new words (underlined) and concepts. After fine-tuning on CLEVR-Humans, a CLEVR-trained model can now reason about obstruction, superlatives, and reflections but still struggles with hypothetical scenarios (rightmost). It also has learned human preference to primarily identify objects by shape (leftmost).}
    \label{fig:CLEVR-Humans-examples}
  \end{figure*}

	\subsection{CLEVR-Humans: Human-Posed Questions} \label{CLEVR-Humans}
    
    	To assess how well visual reasoning models generalize to more realistic, complex, and free-form questions, the CLEVR-Humans dataset was introduced~\cite{IEP}. This dataset contains human-posed questions on CLEVR images along with their corresponding answers. The number of samples is limited --- 18K for training, 7K for validation, and 7K for testing. The questions were collected from Amazon Mechanical Turk workers prompted to ask questions that were likely \textit{hard for a smart robot to answer}. As a result, CLEVR-Humans questions use more diverse vocabulary and complex concepts.
        
        \subsubsection{Method}
		To test FiLM on CLEVR-Humans, we take our best CLEVR-trained FiLM model and fine-tune its FiLM-generating linguistic pipeline alone on CLEVR-Humans. Similar to prior work~\cite{IEP}, we do not update the visual pipeline on CLEVR-Humans to mitigate overfitting to the small training set.
        
        \subsubsection{Results}
		Our model achieves state-of-the-art generalization to CLEVR-Humans, both before and after fine-tuning, as shown in Table~\ref{tab:CLEVR-Humans}, indicating that FiLM is well-suited to handle more complex and diverse questions. Figure~\ref{fig:CLEVR-Humans-examples} shows examples from CLEVR-Humans with FiLM model answers. Before fine-tuning, FiLM outperforms prior methods by a smaller margin. After fine-tuning, FiLM reaches a considerably improved final accuracy. In particular, the \textit{gain} in accuracy made by FiLM upon fine-tuning is more than 50\% greater than those made by other models; FiLM adapts data-efficiently using the small CLEVR-Humans dataset.
        
        Notably, FiLM surpasses the prior state-of-the-art method, Program Generator + Execution Engine (PG+EE), after fine-tuning by \textit{$9.3\%$}. Prior work on PG+EEs explains that this neural module network method struggles on questions which cannot be well approximated with the model's module inventory~\cite{IEP}. In contrast, FiLM has the freedom to modulate existing feature maps, a fairly flexible and fine-grained operation, in novel ways to reason about new concepts. These results thus provide some evidence for the benefits of FiLM's general nature.

    	\begin{table}[t]
          \centering
          {\small
          \begin{tabular}{l|c|cccccc}
          \toprule
          {}      & {Train} & {Train CLEVR,} \\
          {Model} & {CLEVR} & {fine-tune human} \\
          \midrule
          LSTM               & 27.5 & 36.5 \\
          CNN+LSTM           & 37.7 & 43.2 \\
          CNN+LSTM+SA+MLP    & 50.4 & 57.6 \\
          PG+EE (18K prog.)  & 54.0 & 66.6 \\
          \midrule
          CNN+GRU+FiLM & \textbf{56.6} & \textbf{75.9} \\
          \bottomrule
          \end{tabular}
          }
          \caption{\label{tab:CLEVR-Humans} CLEVR-Humans test accuracy, before (left) and after (right) fine-tuning on CLEVR-Humans data}
        \end{table}
    
    \subsection{CLEVR Compositional Generalization Test} 
    	\label{CLEVR-CoGenT}

        To test how well models learn compositional concepts that generalize, CLEVR-CoGenT was introduced~\cite{CLEVR}. This dataset is synthesized in the same way as CLEVR but contains two conditions: in Condition A, all cubes are gray, blue, brown, or yellow and all cylinders are red, green, purple, or cyan; in Condition B, cubes and cylinders swap color palettes. Both conditions contain spheres of all colors. CLEVR-CoGenT thus indicates how a model answers CLEVR questions: by memorizing combinations of traits or by learning disentangled or general representations.
        
\begin{figure}[t]
		\centering
        \includegraphics[width=0.75\linewidth]{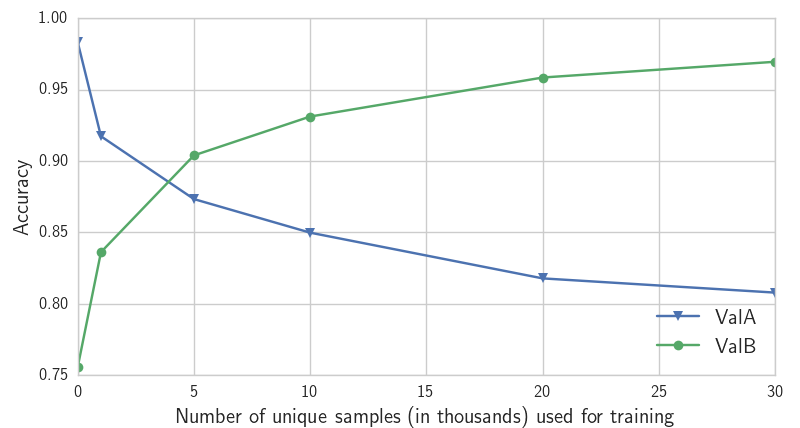}
        \small
  \begin{tabular}{l| c c | c c }
    \toprule
    &  \multicolumn{2}{m{1.8cm}|}{\makecell{\textbf{Train A}}}& \multicolumn{2}{m{1.8cm}}{\makecell{\textbf{Fine-tune B}}}  \\ 
    Method & A & B & A & B \\ \midrule 
    CNN+LSTM+SA & 80.3 & 68.7 & 75.7 & 75.8 \\ 
    PG+EE (18K prog.) & 96.6 & 73.7 & 76.1 & 92.7 \\  \hline
    CNN+GRU+FiLM & \textbf{98.3} & 75.6 & 80.8 & \textbf{96.9} \\ 
    CNN+GRU+FiLM 0-Shot & \textbf{98.3} & \textbf{78.8} & \textbf{81.1} & \textbf{96.9} \\
    \bottomrule 
  \end{tabular}
  \caption{CoGenT results. FiLM ValB accuracy reported on ValB without the 30K fine-tuning samples (Figure). Accuracy before and after fine-tuning on 30K of ValB (Table).}
\label{fig:CoGenT}
\end{figure}
        
        \subsubsection{Results}
          We train our best model architecture on Condition A and report accuracies on Conditions A and B, before and after fine-tuning on B, in Figure~\ref{fig:CoGenT}. Our results indicate FiLM surpasses other visual reasoning models at learning general concepts. FiLM learns better compositional generalization even than PG+EE, which explicitly models compositionality and is trained with program-level supervision that specifically includes filtering colors and filtering shapes.

		\subsubsection{Sample Efficiency and Catastrophic Forgetting} We show sample efficiency and forgetting curves in Figure~\ref{fig:CoGenT}. FiLM achieves prior state-of-the-art accuracy with $1/3$ as much fine-tuning data. However, our FiLM model still suffers from catastrophic forgetting after fine-tuning.
        
		\subsubsection{Zero-Shot Generalization}
        
		FiLM's accuracy on Condition A is much higher than on B, suggesting FiLM has memorized attribute combinations to an extent. For example, the model learns a bias that cubes are not cyan, as learning this training set bias helps minimize training loss.

\begin{figure}[t]
  \centering
  \includegraphics[width=0.45\linewidth]{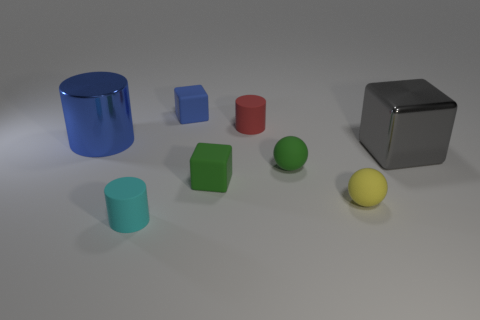}
        \scriptsize
  \begin{tabular}{|l| l |}
    \hline
    Question & What is the blue big cylinder made of? \\ \hline
    (1) Swap shape & What is the blue big \textbf{sphere} made of? \\
	(2) Swap color & What is the \textbf{green} big cylinder made of? \\
	(3) Swap shape/color & What is the \textbf{green} big \textbf{sphere} made of?  \\ \hline
  \end{tabular}
  \caption{A CLEVR-CoGenT example. The combination of concepts \textit{``blue''} and \textit{``cylinder''} is not in the training set. Our zero-shot method computes the original question's FiLM parameters via linear combination of three other questions' FiLM parameters: (1) + (2) - (3). This method corrects our model's answer from \textit{``rubber''} to \textit{``metal''}.}
  \label{fig:zero-shot-generalization}
\end{figure}

    	To overcome this bias, we develop a novel FiLM-based zero-shot generalization method. Inspired by word embedding manipulations, \textit{e.g. ``King'' - ``Man'' + ``Woman'' = ``Queen''}~\cite{mikolov2013distributed}, we test if linear manipulation extends to reasoning with FiLM. We compute $(\bm{\gamma},\bm{\beta})$ for \textit{``How many cyan cubes are there?''} via the linear combination of questions in the FiLM parameter space: \textit{``How many cyan spheres are there?''} $+$ \textit{``How many brown cubes are there?''} $-$ \textit{``How many brown spheres are there?''}. With this $(\bm{\gamma},\bm{\beta})$, our model can correctly count cyan cubes. We show another example of this method in Figure~\ref{fig:zero-shot-generalization}.
    
    We evaluate this method on validation B, using a parser to automatically generate the right combination of questions. We test previously reported CLEVR-CoGenT FiLM models with this method and show results in Figure~\ref{fig:CoGenT}. With this method, there is a $3.2\%$ overall accuracy gain when training on A and testing for zero-shot generalization on B. Yet this method could only be applied to $1/3$ of questions in B. For these questions, model accuracy starts at $71.5\%$ and jumps to $80.7\%$. Before fine-tuning on B, the accuracy between zero-shot and original approaches on A is identical, likewise for B after fine-tuning. We note that difference in the predicted FiLM parameters between these two methods is negligible, likely causing the similar performance.
    
    We achieve these improvements without specifically training our model for zero-shot generalization. Our method simply allows FiLM to take advantage of any concept disentanglement in the CNN after training. We also observe that convex combinations of the FiLM parameters -- \textit{i.e.} between \textit{``How many cyan things are there?''} and \textit{``How many brown things are there?''} -- often monotonically interpolates the predicted answer between the answers to endpoint questions. These results highlight, to a limited extent, the flexibility of FiLM parameters for meaningful manipulations.

    As implemented, this method has many limitations. However, approaches from word embeddings, representation learning, and zero-shot learning can be applied to directly optimize $(\bm{\gamma},\bm{\beta})$ for analogy-making~\cite{TransE,gu2015traversing,46127}. The FiLM-ed network could directly train with this procedure via backpropagation. A learned model could also replace the parser. We find such avenues promising for future work.

\section{Conclusion} \label{conclusion}
	We show that a model can achieve strong visual reasoning using general-purpose Feature-wise Linear Modulation layers. By efficiently manipulating a neural network's intermediate features in a selective and meaningful manner using FiLM layers, a RNN can effectively use language to modulate a CNN to carry out diverse and multi-step reasoning tasks over an image.
    Our ablation study suggests that FiLM is resilient to architectural modifications, test time ablations, and even restrictions on FiLM layers themselves.
    Notably, we provide evidence that FiLM's success is not closely connected with normalization as previously assumed.
    Thus, we open the door for applications of this approach to settings where normalization is less common, such as RNNs and reinforcement learning.
    Our findings also suggest that FiLM models can generalize better, more sample efficiently, and even zero-shot to foreign or more challenging data.
    Overall, the results of our investigation of FiLM in the case of visual reasoning complement broader literature that demonstrates the success of FiLM-like techniques across many domains, supporting the case for FiLM's strength not simply within a single domain but as a general, versatile approach.

\section{Acknowledgements}
We thank the developers of PyTorch (\url{pytorch.org}) and~\cite{IEP} for open-source code which our implementation was based off. We thank Mohammad Pezeshki, Dzmitry Bahdanau, Yoshua Bengio, Nando de Freitas, Hugo Larochelle, Laurens van der Maaten, Joseph Cohen, Joelle Pineau, Olivier Pietquin, J\'er\'emie Mary, C\'esar Laurent, Chin-Wei Huang, Layla Asri, Max Smith, and James Ough for helpful discussions and Justin Johnson for CLEVR test evaluations. We thank NVIDIA for donating a DGX-1 computer used in this work. We also acknowledge FRQNT through the CHIST-ERA IGLU project, Coll\`ege Doctoral Lille Nord de France, and CPER Nord-Pas de Calais/FEDER DATA Advanced data science and technologies 2015-2020 for funding our work. Lastly, we thank~\url{acronymcreator.net} for the acronym FiLM.

\fontsize{9.0pt}{10.0pt} \selectfont
\bibliography{Bibliography-File}

\begin{thebibliography}{}

\bibitem[\protect\citeauthoryear{Anderson \bgroup et al\mbox.\egroup
  }{2017}]{anderson2017bottom}
Anderson, P.; He, X.; Buehler, C.; Teney, D.; Johnson, M.; Gould, S.; and
  Zhang, L.
\newblock 2017.
\newblock Bottom-up and top-down attention for image captioning and vqa.
\newblock In {\em VQA Workshop at CVPR}.

\bibitem[\protect\citeauthoryear{Andreas \bgroup et al\mbox.\egroup
  }{2016a}]{NMNQA}
Andreas, J.; Marcus, R.; Darrell, T.; and Klein, D.
\newblock 2016a.
\newblock Learning to compose neural networks for question answering.
\newblock In {\em NAACL}.

\bibitem[\protect\citeauthoryear{Andreas \bgroup et al\mbox.\egroup
  }{2016b}]{NMN}
Andreas, J.; Rohrbach, M.; Darrell, T.; and Klein, D.
\newblock 2016b.
\newblock Neural module networks.
\newblock In {\em CVPR}.

\bibitem[\protect\citeauthoryear{Antol \bgroup et al\mbox.\egroup
  }{2015}]{antol2015}
Antol, S.; Agrawal, A.; Lu, J.; Mitchell, M.; Batra, D.; Zitnick, C.~L.; and
  Parikh, D.
\newblock 2015.
\newblock {VQA}: {V}isual {Q}uestion {A}nswering.
\newblock In {\em ICCV}.

\bibitem[\protect\citeauthoryear{Bordes \bgroup et al\mbox.\egroup
  }{2013}]{TransE}
Bordes, A.; Usunier, N.; Garcia-Duran, A.; Weston, J.; and Yakhnenko, O.
\newblock 2013.
\newblock Translating embeddings for modeling multi-relational data.
\newblock In Burges, C. J.~C.; Bottou, L.; Welling, M.; Ghahramani, Z.; and
  Weinberger, K.~Q., eds., {\em NIPS}. Curran Associates, Inc.
\newblock  2787--2795.

\bibitem[\protect\citeauthoryear{Chung \bgroup et al\mbox.\egroup }{2014}]{GRU}
Chung, J.; G{\"{u}}l{\c{c}}ehre, {\c{C}}.; Cho, K.; and Bengio, Y.
\newblock 2014.
\newblock Empirical evaluation of gated recurrent neural networks on sequence
  modeling.
\newblock In {\em Deep Learning Workshop at NIPS}.

\bibitem[\protect\citeauthoryear{de Vries \bgroup et al\mbox.\egroup
  }{2017}]{modulating_vision}
de~Vries, H.; Strub, F.; Mary, J.; Larochelle, H.; Pietquin, O.; and Courville,
  A.~C.
\newblock 2017.
\newblock Modulating early visual processing by language.
\newblock In {\em NIPS}.

\bibitem[\protect\citeauthoryear{Dumoulin, Shlens, and Kudlur}{2017}]{CIN}
Dumoulin, V.; Shlens, J.; and Kudlur, M.
\newblock 2017.
\newblock A learned representation for artistic style.
\newblock In {\em ICLR}.

\bibitem[\protect\citeauthoryear{Eigen, Ranzato, and Sutskever}{2014}]{DeepMoE}
Eigen, D.; Ranzato, M.; and Sutskever, I.
\newblock 2014.
\newblock Learning factored representations in a deep mixture of experts.
\newblock In {\em ICLR Workshops}.

\bibitem[\protect\citeauthoryear{Gehring \bgroup et al\mbox.\egroup
  }{2017}]{pmlr-v70-gehring17a}
Gehring, J.; Auli, M.; Grangier, D.; Yarats, D.; and Dauphin, Y.~N.
\newblock 2017.
\newblock Convolutional sequence to sequence learning.
\newblock In {\em ICML}.

\bibitem[\protect\citeauthoryear{Geman \bgroup et al\mbox.\egroup
  }{2015}]{geman2015visual}
Geman, D.; Geman, S.; Hallonquist, N.; and Younes, L.
\newblock 2015.
\newblock {Visual turing test for computer vision systems}.
\newblock volume 112,  3618--3623.
\newblock National Acad Sciences.

\bibitem[\protect\citeauthoryear{Ghiasi \bgroup et al\mbox.\egroup
  }{2017}]{CIN2}
Ghiasi, G.; Lee, H.; Kudlur, M.; Dumoulin, V.; and Shlens, J.
\newblock 2017.
\newblock Exploring the structure of a real-time, arbitrary neural artistic
  stylization network.
\newblock {\em CoRR abs/1705.06830}.

\bibitem[\protect\citeauthoryear{Goyal \bgroup et al\mbox.\egroup
  }{2017}]{goyal2016making}
Goyal, Y.; Khot, T.; Summers{-}Stay, D.; Batra, D.; and Parikh, D.
\newblock 2017.
\newblock Making the {V} in {VQA} matter: Elevating the role of image
  understanding in {V}isual {Q}uestion {A}nswering.
\newblock In {\em CVPR}.

\bibitem[\protect\citeauthoryear{Guu, Miller, and
  Liang}{2015}]{gu2015traversing}
Guu, K.; Miller, J.; and Liang, P.
\newblock 2015.
\newblock Traversing knowledge graphs in vector space.
\newblock In {\em EMNLP}.

\bibitem[\protect\citeauthoryear{Ha, Dai, and Le}{2016}]{Hypernets}
Ha, D.; Dai, A.; and Le, Q.
\newblock 2016.
\newblock Hypernetworks.
\newblock In {\em ICLR}.

\bibitem[\protect\citeauthoryear{He \bgroup et al\mbox.\egroup }{2016}]{ResNet}
He, K.; Zhang, X.; Ren, S.; and Sun, J.
\newblock 2016.
\newblock Deep residual learning for image recognition.
\newblock In {\em CVPR}.

\bibitem[\protect\citeauthoryear{Hochreiter and Schmidhuber}{1997}]{LSTM}
Hochreiter, S., and Schmidhuber, J.
\newblock 1997.
\newblock Long short-term memory.
\newblock {\em Neural Comput.} 9(8):1735--1780.

\bibitem[\protect\citeauthoryear{Hu \bgroup et al\mbox.\egroup }{2017}]{N2NMN}
Hu, R.; Andreas, J.; Rohrbach, M.; Darrell, T.; and Saenko, K.
\newblock 2017.
\newblock Learning to reason: End-to-end module networks for visual question
  answering.
\newblock In {\em ICCV}.

\bibitem[\protect\citeauthoryear{Hu, Shen, and Sun}{2017}]{hu2017}
Hu, J.; Shen, L.; and Sun, G.
\newblock 2017.
\newblock {Squeeze-and-Excitation Networks}.
\newblock In {\em ILSVRC 2017 Workshop at CVPR}.

\bibitem[\protect\citeauthoryear{Huang and Belongie}{2017}]{AIN}
Huang, X., and Belongie, S.
\newblock 2017.
\newblock Arbitrary style transfer in real-time with adaptive instance
  normalization.
\newblock In {\em ICCV}.

\bibitem[\protect\citeauthoryear{Ioffe and Szegedy}{2015}]{BN}
Ioffe, S., and Szegedy, C.
\newblock 2015.
\newblock Batch normalization: Accelerating deep network training by reducing
  internal covariate shift.
\newblock In {\em ICML}.

\bibitem[\protect\citeauthoryear{Johnson \bgroup et al\mbox.\egroup
  }{2017a}]{CLEVR}
Johnson, J.; Hariharan, B.; van~der Maaten, L.; Fei{-}Fei, L.; Zitnick, C.~L.;
  and Girshick, R.~B.
\newblock 2017a.
\newblock {CLEVR:} {A} diagnostic dataset for compositional language and
  elementary visual reasoning.
\newblock In {\em CVPR}.

\bibitem[\protect\citeauthoryear{Johnson \bgroup et al\mbox.\egroup
  }{2017b}]{IEP}
Johnson, J.; Hariharan, B.; van~der Maaten, L.; Hoffman, J.; Li, F.; Zitnick,
  C.~L.; and Girshick, R.~B.
\newblock 2017b.
\newblock Inferring and executing programs for visual reasoning.
\newblock In {\em ICCV}.

\bibitem[\protect\citeauthoryear{Jordan and Jacobs}{1994}]{HME}
Jordan, M.~I., and Jacobs, R.~A.
\newblock 1994.
\newblock Hierarchical mixtures of experts and the em algorithm.
\newblock {\em Neural Comput.} 6(2):181--214.

\bibitem[\protect\citeauthoryear{Kim, Song, and
  Bengio}{2017}]{DynamicLayerNorm}
Kim, T.; Song, I.; and Bengio, Y.
\newblock 2017.
\newblock Dynamic layer normalization for adaptive neural acoustic modeling in
  speech recognition.
\newblock In {\em InterSpeech}.

\bibitem[\protect\citeauthoryear{Kingma and Ba}{2015}]{Adam}
Kingma, D.~P., and Ba, J.
\newblock 2015.
\newblock Adam: {A} method for stochastic optimization.
\newblock In {\em ICLR}.

\bibitem[\protect\citeauthoryear{Kirkpatrick \bgroup et al\mbox.\egroup
  }{2017}]{OCF}
Kirkpatrick, J.; Pascanu, R.; Rabinowitz, N.; Veness, J.; Desjardins, G.; Rusu,
  A.~A.; Milan, K.; Quan, J.; Ramalho, T.; Grabska-Barwinska, A.; Hassabis, D.;
  Clopath, C.; Kumaran, D.; and Hadsell, R.
\newblock 2017.
\newblock Overcoming catastrophic forgetting in neural networks.
\newblock {\em National Academy of Sciences} 114(13):3521--3526.

\bibitem[\protect\citeauthoryear{Lu \bgroup et al\mbox.\egroup
  }{2016}]{lu2016hierarchical}
Lu, J.; Yang, J.; Batra, D.; and Parikh, D.
\newblock 2016.
\newblock Hierarchical question-image co-attention for visual question
  answering.
\newblock In {\em NIPS}.

\bibitem[\protect\citeauthoryear{Malinowski and
  Fritz}{2014}]{malinowski2014multi}
Malinowski, M., and Fritz, M.
\newblock 2014.
\newblock {A multi-world approach to question answering about real-world scenes
  based on uncertain input}.
\newblock In {\em NIPS}.

\bibitem[\protect\citeauthoryear{Malinowski, Rohrbach, and
  Fritz}{2015}]{malinowski2015ask}
Malinowski, M.; Rohrbach, M.; and Fritz, M.
\newblock 2015.
\newblock Ask your neurons: A neural-based approach to answering questions
  about images.
\newblock In {\em ICCV}.

\bibitem[\protect\citeauthoryear{Mikolov \bgroup et al\mbox.\egroup
  }{2013}]{mikolov2013distributed}
Mikolov, T.; Sutskever, I.; Chen, K.; Corrado, G.~S.; and Dean, J.
\newblock 2013.
\newblock Distributed representations of words and phrases and their
  compositionality.
\newblock In {\em NIPS}.

\bibitem[\protect\citeauthoryear{Oh \bgroup et al\mbox.\egroup }{2017}]{46127}
Oh, J.; Singh, S.; Lee, H.; and Kholi, P.
\newblock 2017.
\newblock Zero-shot task generalization with multi-task deep reinforcement
  learning.
\newblock In {\em ICML}.

\bibitem[\protect\citeauthoryear{Perez \bgroup et al\mbox.\egroup
  }{2017}]{LVRWSP}
Perez, E.; de~Vries, H.; Strub, F.; Dumoulin, V.; and Courville, A.~C.
\newblock 2017.
\newblock Learning visual reasoning without strong priors.
\newblock In {\em MLSLP Workshop at ICML}.

\bibitem[\protect\citeauthoryear{Radford, Metz, and Chintala}{2016}]{DCGAN}
Radford, A.; Metz, L.; and Chintala, S.
\newblock 2016.
\newblock Unsupervised representation learning with deep convolutional
  generative adversarial networks.
\newblock In {\em ICLR}.

\bibitem[\protect\citeauthoryear{Russakovsky \bgroup et al\mbox.\egroup
  }{2015}]{ImageNet}
Russakovsky, O.; Deng, J.; Su, H.; Krause, J.; Satheesh, S.; Ma, S.; Huang, Z.;
  Karpathy, A.; Khosla, A.; Bernstein, M.~S.; Berg, A.~C.; and Li, F.
\newblock 2015.
\newblock Imagenet large scale visual recognition challenge.
\newblock {\em IJCV} 115(3):211--252.

\bibitem[\protect\citeauthoryear{Santoro \bgroup et al\mbox.\egroup
  }{2017}]{RN}
Santoro, A.; Raposo, D.; Barrett, D.~G.; Malinowski, M.; Pascanu, R.;
  Battaglia, P.; and Lillicrap, T.
\newblock 2017.
\newblock A simple neural network module for relational reasoning.
\newblock {\em CoRR abs/1706.01427}.

\bibitem[\protect\citeauthoryear{Shazeer \bgroup et al\mbox.\egroup
  }{2017}]{shazeer2017outrageously}
Shazeer, N.; Mirhoseini, A.; Maziarz, K.; Davis, A.; Le, Q.; Hinton, G.; and
  Dean, J.
\newblock 2017.
\newblock Outrageously large neural networks: The sparsely-gated
  mixture-of-experts layer.
\newblock In {\em ICLR}.

\bibitem[\protect\citeauthoryear{van~den Oord \bgroup et al\mbox.\egroup
  }{2016a}]{WaveNet}
van~den Oord, A.; Dieleman, S.; Zen, H.; Simonyan, K.; Vinyals, O.; Graves, A.;
  Kalchbrenner, N.; Senior, A.; and Kavukcuoglu, K.
\newblock 2016a.
\newblock Wavenet: A generative model for raw audio.
\newblock {\em CoRR abs/1609.03499}.

\bibitem[\protect\citeauthoryear{van~den Oord \bgroup et al\mbox.\egroup
  }{2016b}]{van2016conditional}
van~den Oord, A.; Kalchbrenner, N.; Espeholt, L.; Vinyals, O.; Graves, A.; and
  Kavukcuoglu, K.
\newblock 2016b.
\newblock Conditional image generation with pixelcnn decoders.
\newblock In {\em NIPS}.

\bibitem[\protect\citeauthoryear{van~der Maaten and
  Hinton}{2008}]{maaten2008visualizing}
van~der Maaten, L., and Hinton, G.
\newblock 2008.
\newblock Visualizing data using t-sne.
\newblock {\em JMLR} 9(Nov):2579--2605.

\bibitem[\protect\citeauthoryear{Watters \bgroup et al\mbox.\egroup
  }{2017}]{VIN}
Watters, N.; Tacchetti, A.; Weber, T.; Pascanu, R.; Battaglia, P.; and Zoran,
  D.
\newblock 2017.
\newblock Visual interaction networks.
\newblock {\em CoRR abs/1706.01433}.

\bibitem[\protect\citeauthoryear{Yang \bgroup et al\mbox.\egroup }{2016}]{SANs}
Yang, Z.; He, X.; Gao, J.; Deng, L.; and Smola, A.~J.
\newblock 2016.
\newblock Stacked attention networks for image question answering.
\newblock In {\em CVPR}.

\end{thebibliography}
\bibliographystyle{aaai}
\section{Appendix}
  
	\subsection{Error Analysis} \label{errors}

\begin{figure}[th!]
   	\small{
      \begin{tabular}{m{3.7cm}m{3.7cm}}
  	  \textbf{Q:} \textit{Is there a big brown object of the same shape as the green thing?} \textbf{A:} \textit{Yes} (\textbf{P:} \textit{No})
  	  & 
      \textbf{Q:} \textit{What number of other things are the same material as the big gray cylinder?} \textbf{A:} \textit{6} (\textbf{P:} \textit{5})
      \\
      \includegraphics[width=\linewidth]{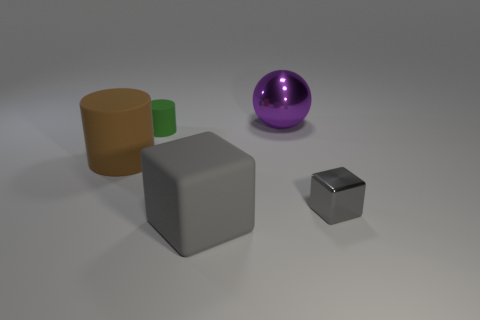}
	  & 	
  	  \includegraphics[width=\linewidth]{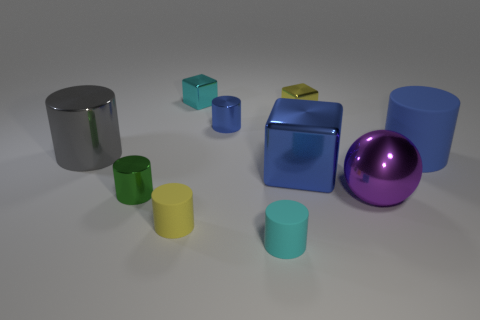}
      \\
      \includegraphics[width=\linewidth]{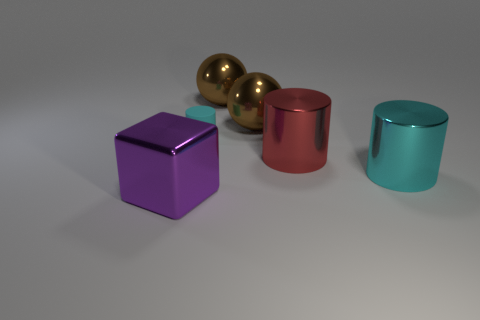}
	  &
  	  \includegraphics[width=\linewidth]{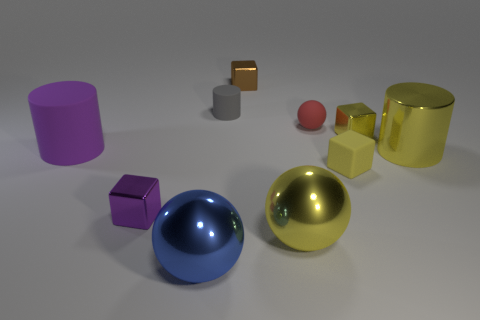}
	  \cr
	  \\
      \textbf{Q:} \textit{What shape is the big metal thing that is the same color as the small cylinder?} \textbf{A:} \textit{Cylinder} (\textbf{P:} \textit{Sphere})
      &
      \textbf{Q:} \textit{How many other things are the same material as the tiny sphere?} \textbf{A:} \textit{3} (\textbf{P:} \textit{2})
	  \end{tabular}
    }
    \caption{Some image-question pairs where our model predicts incorrectly. Most errors we observe are due to partially occluded objects, as highlighted in the three first examples.}
    \label{fig:errors}
\end{figure}

\begin{figure}[th!]
    \centering
    \includegraphics[width=0.7\linewidth]{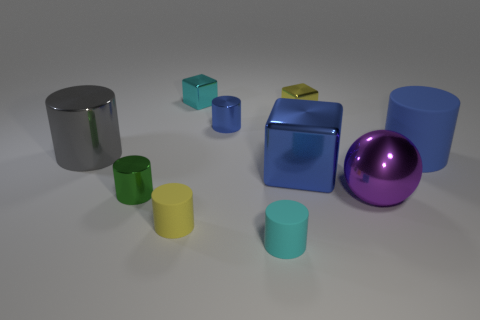}
    \scriptsize
    \begin{tabular}{|l|l|}
    \hline
    Question & Answer\\
    \hline
    How many gray things are there? & 1\\
    How many cyan things are there? & 2\\
    Are there as many gray things as cyan things? & \bf{Yes}\\
    Are there more gray things than cyan things? & No\\
    Are there fewer gray things than cyan things? & Yes\\
    \hline
    \end{tabular}
    \caption{An interesting failure example where our model counts correctly but compares counts erroneously. Its third answer is incorrect and inconsistent with its other answers.}
    \label{fig:logic-error}
\end{figure}

        We examine the errors our model makes to understand where our model fails and how it acts when it does. Examples of these errors are shown in Figures~\ref{fig:errors} and~\ref{fig:logic-error}.
        
        \subsubsection{Occlusion}
        Many model errors are due to partial occlusion. These errors may likely be fixed using a CNN that operates at a higher resolution, which is feasible since FiLM has a computational cost that is independent of resolution.
        
        \subsubsection{Counting}
		$96.1\%$ of counting mistakes are off-by-one errors, showing FiLM has learned underlying concepts behind counting such as close relationships between close numbers.

		\subsubsection{Logical Consistency}
    	The model sometimes makes curious reasoning mistakes a human would not. For example, we find a case where our model correctly counts one gray object and two cyan objects but simultaneously answers that there are the same number of gray and cyan objects. In fact, it answers that the number of gray objects is both less than \textit{and} equal to the number of yellow blocks. These errors could be prevented by directly minimizing logical inconsistency, an interesting avenue for future work orthogonal to FiLM.
    
    \subsection{Model Details}
    Rather than output $\gamma_{i,c}$ directly, we output $\Delta\gamma_{i,c}$, where:
\begin{equation}
	\gamma_{i,c}=1+\Delta\gamma_{i,c},
\end{equation}
since initially zero-centered $\gamma_{i,c}$ can zero out CNN feature map activations and thus gradients. In our implementation, we opt to output $\Delta\gamma_{i,c}$ rather than $\gamma_{i,c}$, but for simplicity, throughout our paper, we explain FiLM using $\gamma_{i,c}$. However, this modification does not seem to affect our model's performance on CLEVR statistically significantly.

\begin{figure}[hb]
	\centering
	\includegraphics[width=0.75\linewidth]{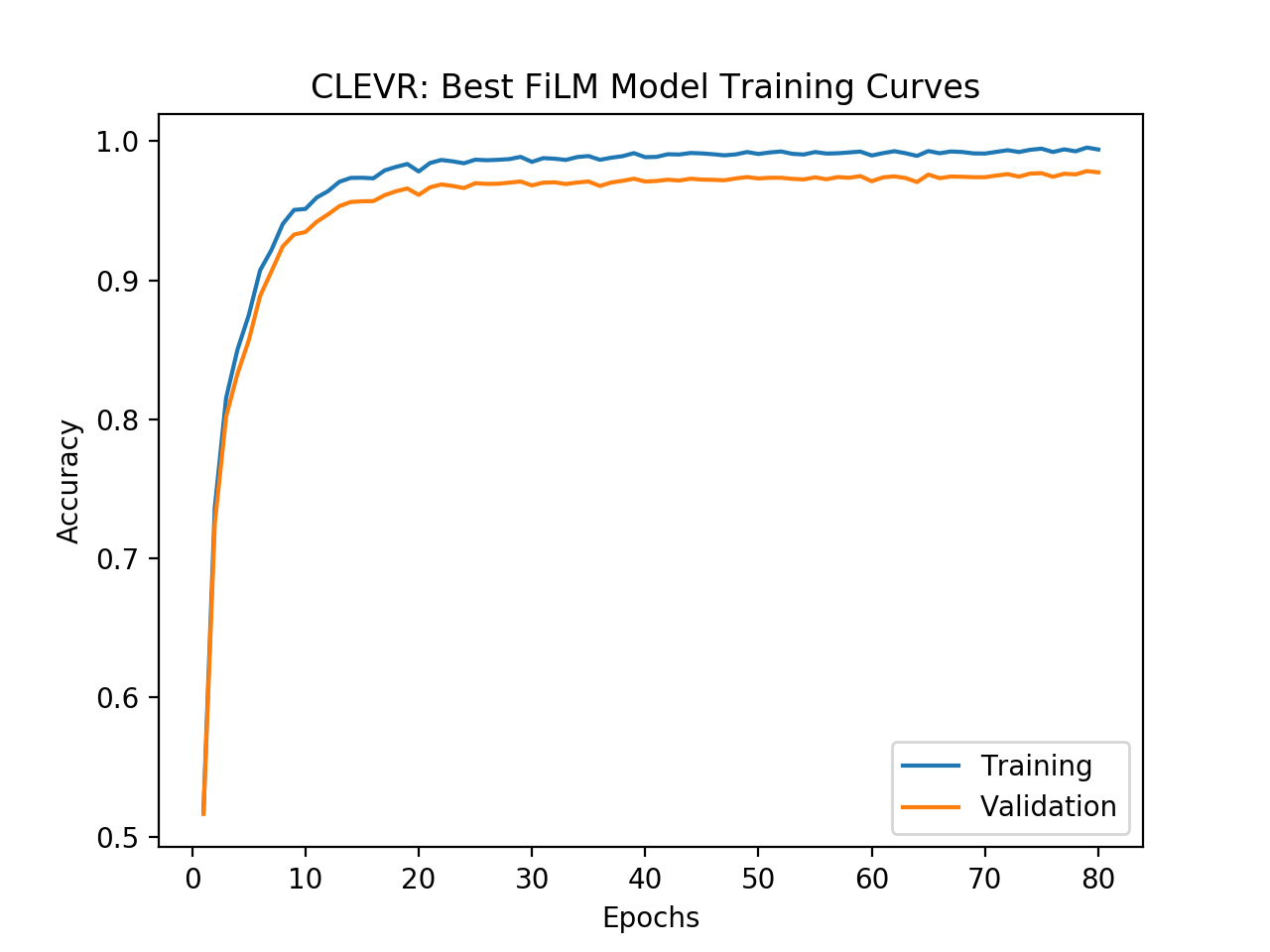}
	\caption{Best model training and validation curves.}
    \label{fig:training-curves}
\end{figure}

	We present training and validation curves for best model trained from image features in Figure~\ref{fig:training-curves}. We observe fast accuracy gains initially, followed by slow, steady increases to a best validation accuracy of $97.84\%$, at which point training accuracy is $99.53\%$. We train on CLEVR for 80 epochs, which takes 4 days using 1 NVIDIA TITAN Xp GPU when learning from image features. For practical reasons, we stop training on CLEVR after 80 epochs, but we observe that accuracy continues to increase slowly even afterwards.
    
\subsection{What Do FiLM Layers Learn?} \label{appendix-what-is-learned}
\begin{figure*}[th]
   \small{
           \begin{tabular}{ccm{2.85cm}m{2.85cm}m{2.85cm}m{2.85cm}}
         \multirow{1}{*}{
         \makecell{
 		\includegraphics[width=0.2\linewidth]{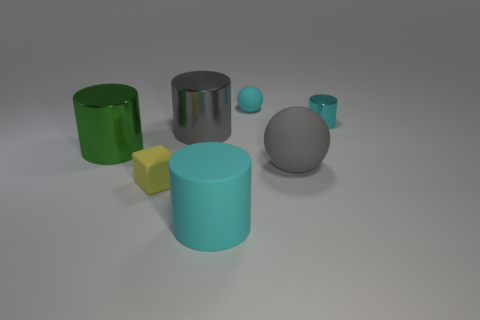} \\ Feature 14 - Block 1
         }}
         & 
         \rotatebox[origin=c]{90}{Before FiLM}
         &
         \includegraphics[width=2.85cm]{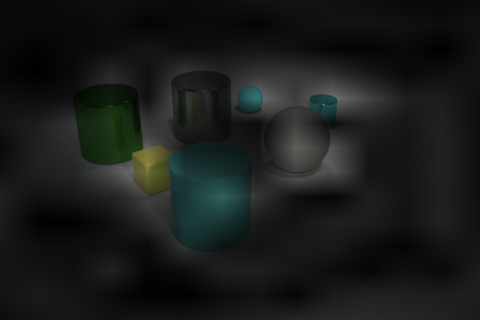}
         & 
         \includegraphics[width=2.85cm]{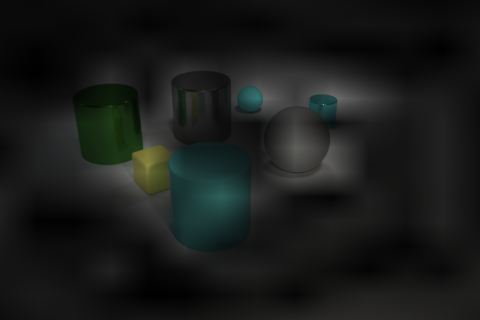}
         & 
   		\includegraphics[width=2.85cm]{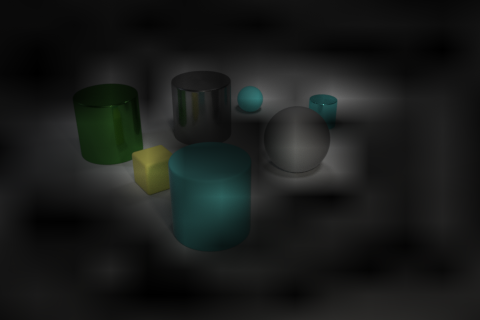}
         & 
         \includegraphics[width=2.85cm]{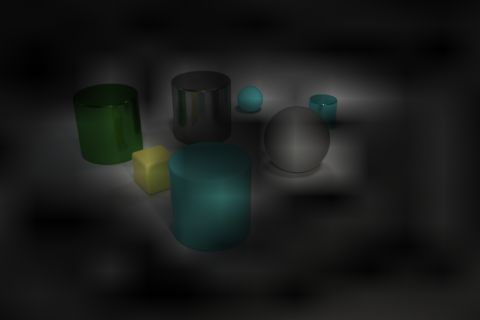}
         \\
         &        
         \rotatebox[origin=c]{90}{After FiLM}
         &
         \includegraphics[width=2.85cm]{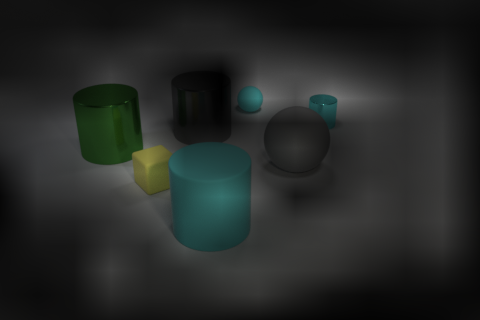}
         & 
         \includegraphics[width=2.85cm]{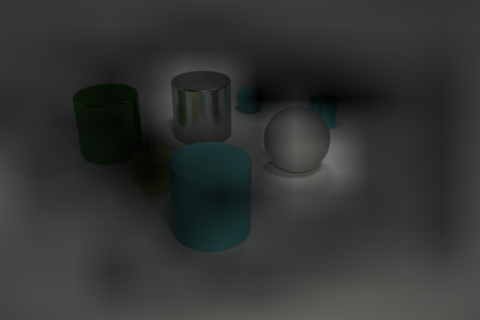}
         & 
   		\includegraphics[width=2.85cm]{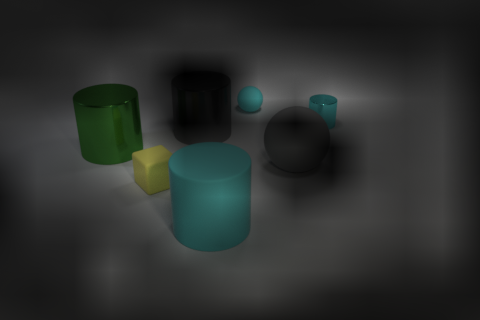}
         &
         \includegraphics[width=2.85cm]{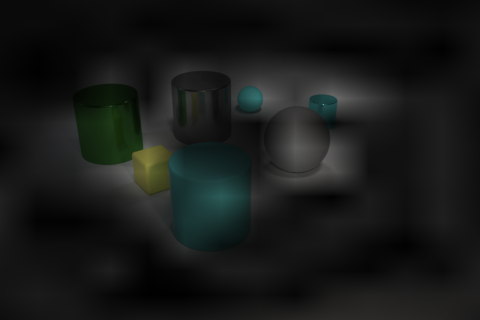}
         \\
         &
         &
         \textbf{Q:} \textit{What is the color of the large rubber cylinder?} \textbf{A:} \textit{Cyan}
         & 
         \textbf{Q:} \textit{What is the color of the large rubber sphere?} \textbf{A:} \textit{Gray}
         & 
   		\textbf{Q:} \textit{What is the color of the cube?} \textbf{A:} \textit{Yellow}
         &
         \textbf{Q:} \textit{How many cylinders are there?}  \textbf{A:} \textit{4}        
\\
&&&&&
\\
         \multirow{1}{*}{
         \makecell{
 		\includegraphics[width=0.2\linewidth]{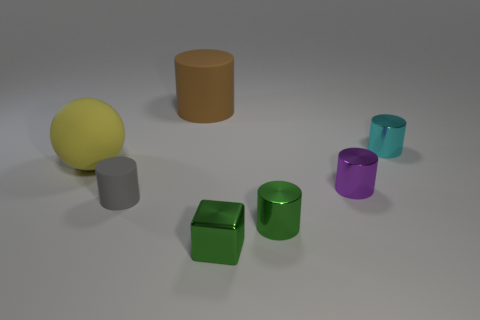}\\ Feature 14 - Block 1
         }}
         &
         \rotatebox[origin=c]{90}{Before FiLM}
         &
         \includegraphics[width=2.85cm]{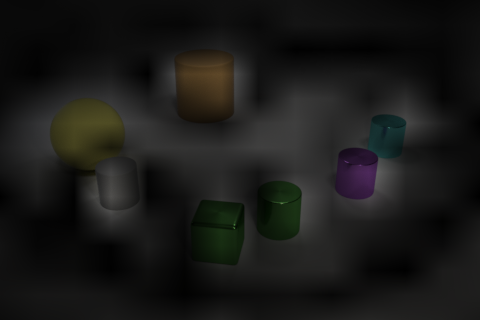}
         & 
         \includegraphics[width=2.85cm]{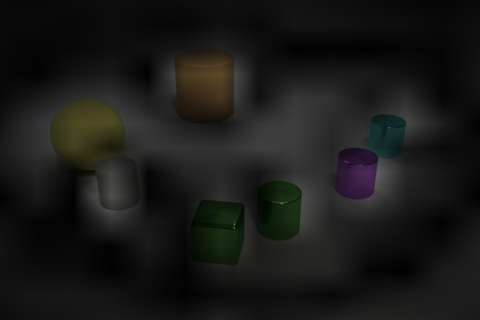}
         & 
   		\includegraphics[width=2.85cm]{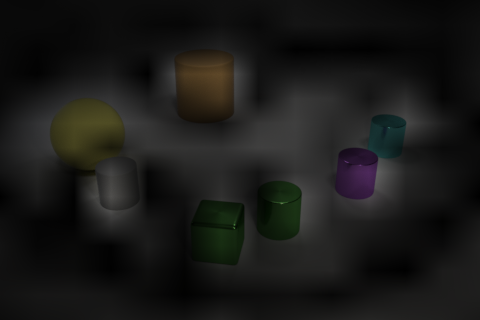}
         & 
         \includegraphics[width=2.85cm]{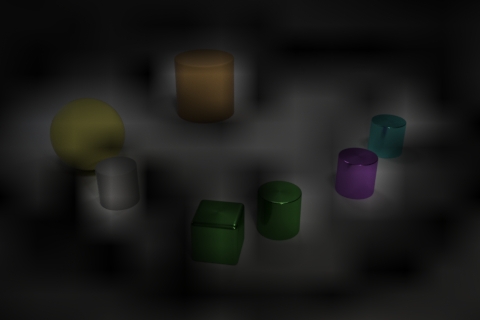}
         \\
         &        
         \rotatebox[origin=c]{90}{After FiLM}
         &
         \includegraphics[width=2.85cm]{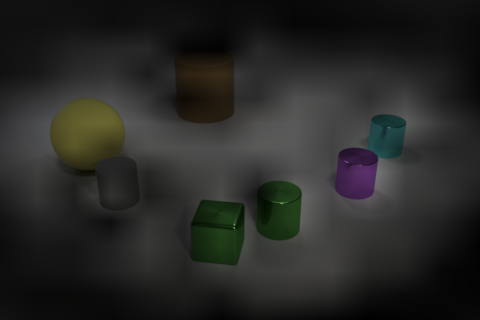}
         & 
         \includegraphics[width=2.85cm]{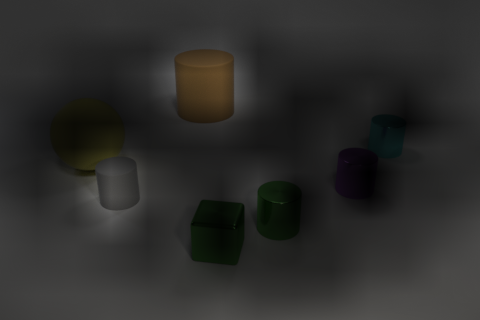}
         &  
   		\includegraphics[width=2.85cm]{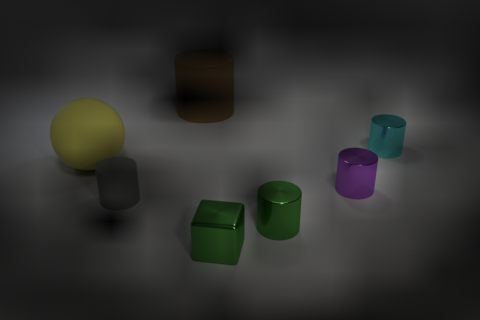}
         &
         \includegraphics[width=2.85cm]{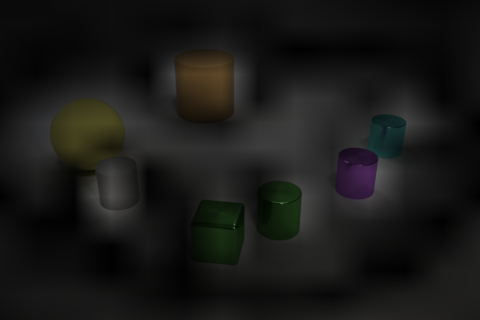}
         \\
         &
         &
         \textbf{Q:} \textit{What is the color of the large rubber cylinder?} \textbf{A:} \textit{Yellow}
         & 
         \textbf{Q:} \textit{What is the color of the large rubber sphere?} \textbf{A:} \textit{Gray}
         & 
   	 	\textbf{Q:} \textit{What is the color of the cube?} \textbf{A:} \textit{Yellow}
         &
         \textbf{Q:} \textit{How many cylinders are there?}  \textbf{A:} \textit{4}        
 	\end{tabular}
    }
      \caption{Visualizations of feature map activations (scaled from 0 to 1) before and after FiLM for a \textit{single} arbitrary feature map from the first ResBlock. This particular feature map seems to detect gray and brown colors. Interestingly, FiLM modifies activations for specifically colored objects for color-specific questions but leaves activations alone for color-agnostic questions. Note that since this is the first FiLM layer, pre-FiLM activations (Rows 1 and 3) for all questions are identical, and differences in post-FiLM activations (Rows 2 and 4) are solely due FiLM's use of question information.}
      \label{fig:film-effect-1}
\end{figure*}

\begin{figure*}[th]
     	\small{
           \begin{tabular}{ccm{2.85cm}m{2.85cm}m{2.85cm}m{2.85cm}}
         \multirow{1}{*}{
         \makecell{
 		\includegraphics[width=0.2\linewidth]{Img25/CLEVR_val_000025.png} \\ Feature 79 - Block 4
         }}
         & 
         \rotatebox[origin=c]{90}{Before FiLM}
         &
         \includegraphics[width=2.85cm]{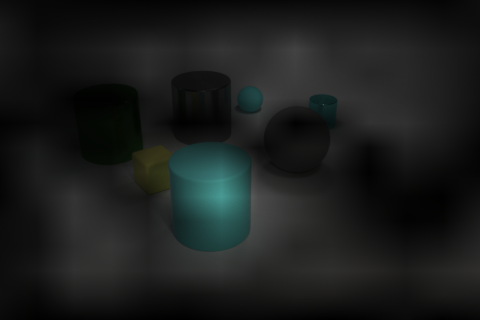}
         & 
         \includegraphics[width=2.85cm]{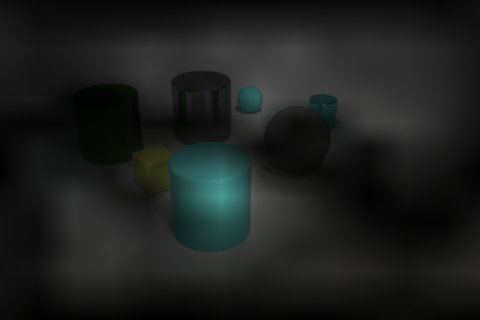}
         & 
   		\includegraphics[width=2.85cm]{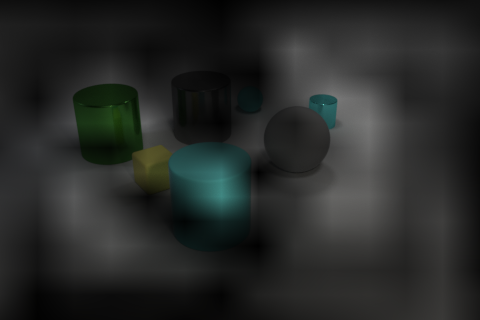}
         & 
         \includegraphics[width=2.85cm]{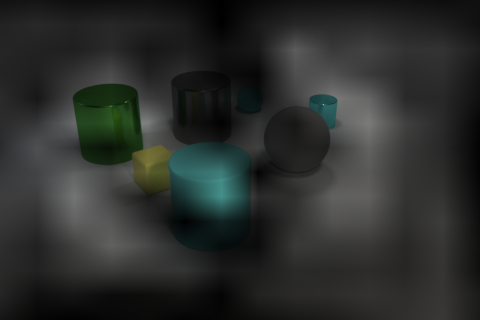}
         \\
         &        
         \rotatebox[origin=c]{90}{After FiLM}
         &
         \includegraphics[width=2.85cm]{Img25/b-1-0-14-after.png}
         & 
         \includegraphics[width=2.85cm]{Img25/b-2-0-14-after.png}
         & 
   		\includegraphics[width=2.85cm]{Img25/b-3-0-14-after.png}
         &
         \includegraphics[width=2.85cm]{Img25/b-4-0-14-after.png}
         \\
         &
         &
         \textbf{Q:} \textit{How many cyan objects are behind the gray sphere?} \textbf{A:} \textit{2}
         & 
         \textbf{Q:} \textit{How many cyan objects are in front of the gray sphere?} \textbf{A:} \textit{1}
         & 
   		\textbf{Q:} \textit{How many cyan objects are left of the gray sphere?} \textbf{A:} \textit{2}
         &
         \textbf{Q:} \textit{How many cyan objects are right of the gray sphere?}  \textbf{A:} \textit{1}        
 	\end{tabular}
    }
      \caption{Visualization of the impact of FiLM for a \textit{single} arbitrary feature map from the last ResBlock. This particular feature map seems to focus on spatial features (\textit{i.e.} front/back or left/right) Note that since this is the last FiLM layer, the top row activations have already been influenced by question information via several FiLM layers.}
      \label{fig:film-effect-2}
\end{figure*}

\begin{figure*}[ht]
        \begin{subfigure}[b]{0.25\textwidth}
                \includegraphics[width=\linewidth]{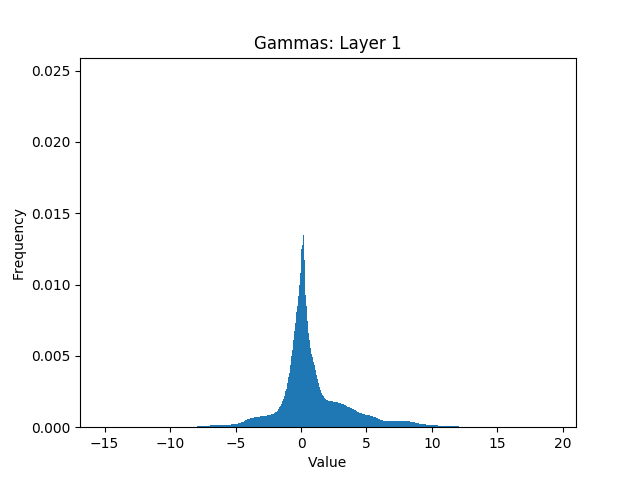}
        \end{subfigure}%
        \begin{subfigure}[b]{0.25\textwidth}
                \includegraphics[width=\linewidth]{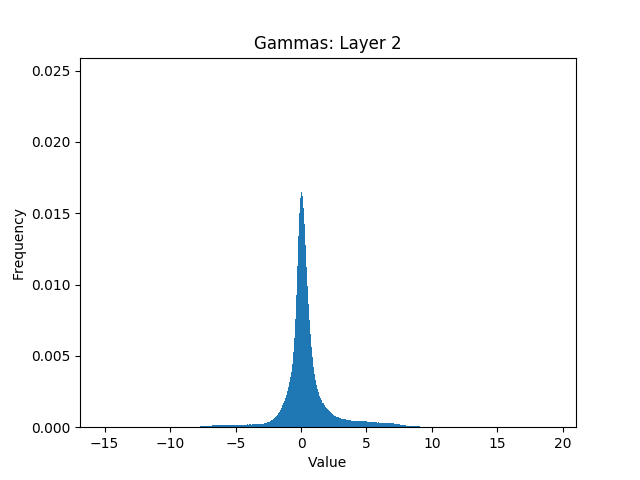}
        \end{subfigure}%
        \begin{subfigure}[b]{0.25\textwidth}
                \includegraphics[width=\linewidth]{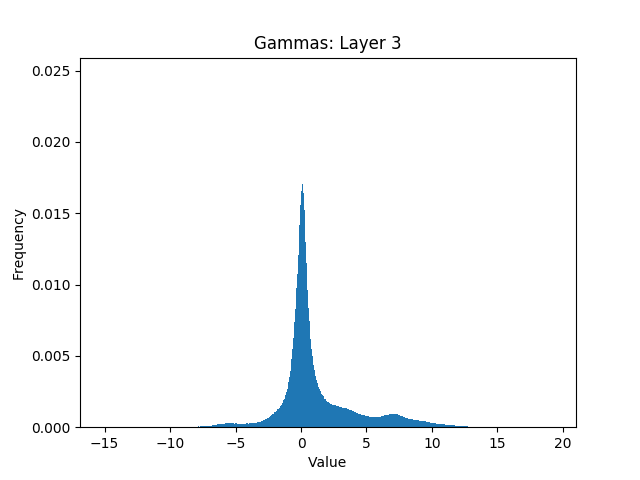}
        \end{subfigure}%
        \begin{subfigure}[b]{0.25\textwidth}
                \includegraphics[width=\linewidth]{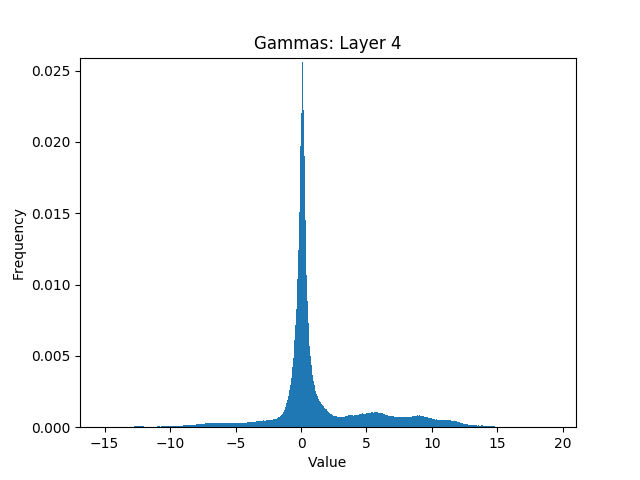}
        \end{subfigure}
        \caption{Histograms of $\gamma_{i,c}$ values for each FiLM layer (layers 1-4 from left to right), computed on CLEVR's validation set. Plots are scaled identically. FiLM layers appear gradually more selective and higher variance.}
        \label{fig:gammas-layerwise}
\end{figure*}

\begin{figure*}[ht]
        \begin{subfigure}[b]{0.25\textwidth}
                \includegraphics[width=\linewidth]{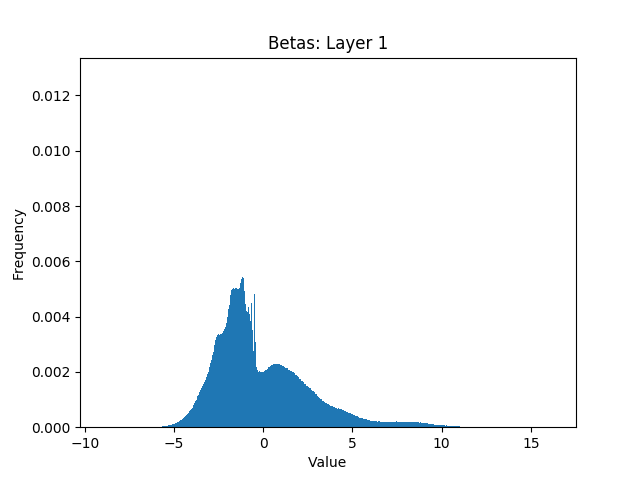}
        \end{subfigure}%
        \begin{subfigure}[b]{0.25\textwidth}
                \includegraphics[width=\linewidth]{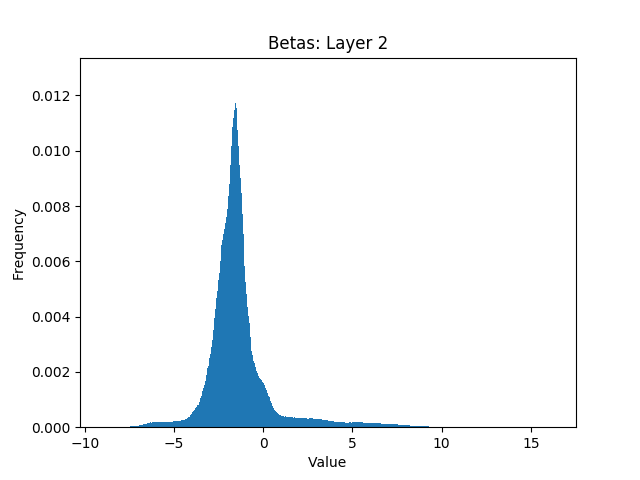}
        \end{subfigure}%
        \begin{subfigure}[b]{0.25\textwidth}
                \includegraphics[width=\linewidth]{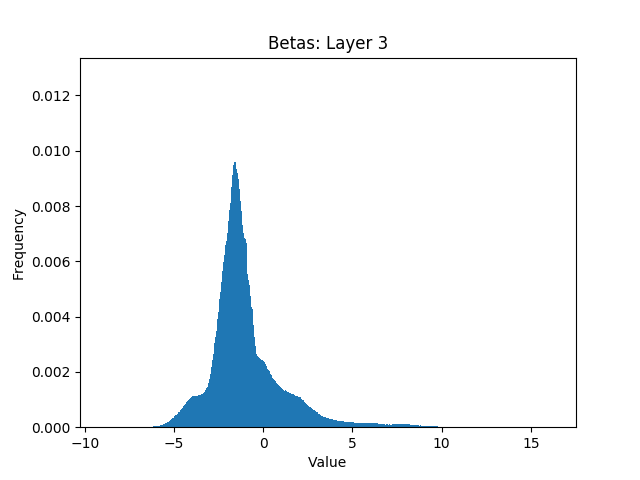}
        \end{subfigure}%
        \begin{subfigure}[b]{0.25\textwidth}
                \includegraphics[width=\linewidth]{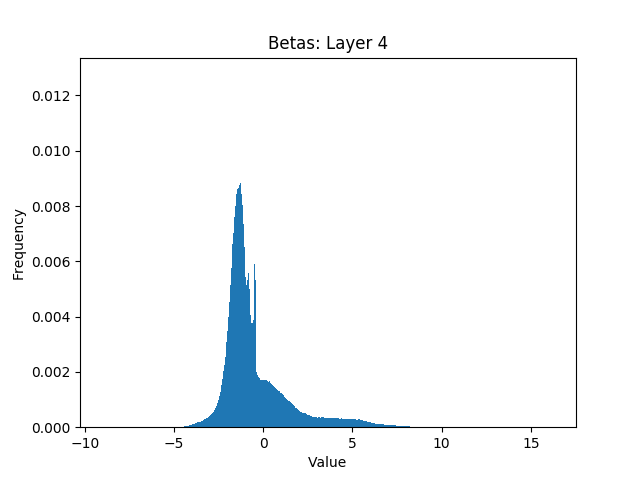}
        \end{subfigure}
        \caption{Histograms of $\beta_{i,c}$ values for each FiLM layer (layers 1-4 from left to right) computed on CLEVR's validation set. Plots are scaled identically. $\beta_{i,c}$ values take a different, higher variance distribution in the first layer than in later layers.}
        \label{fig:betas-layerwise}
\end{figure*}

\begin{figure*}[ht]
        \begin{subfigure}[b]{0.25\textwidth}
                \includegraphics[width=\linewidth]{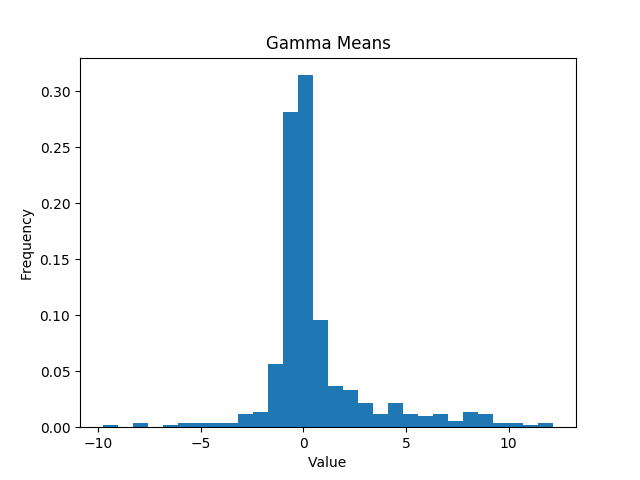}
        \end{subfigure}%
        \begin{subfigure}[b]{0.25\textwidth}
                \includegraphics[width=\linewidth]{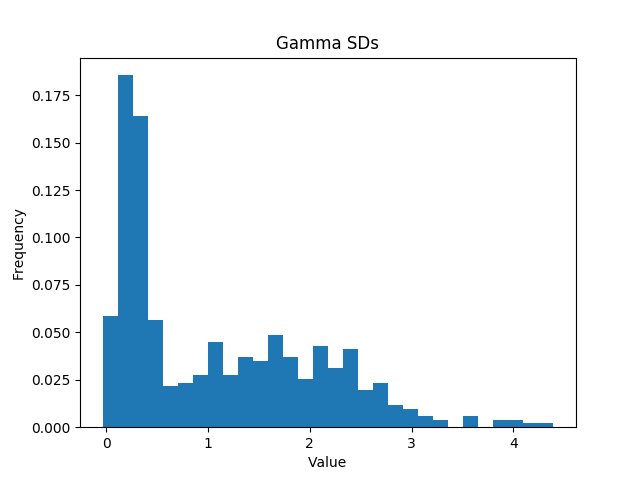}
        \end{subfigure}%
        \begin{subfigure}[b]{0.25\textwidth}
                \includegraphics[width=\linewidth]{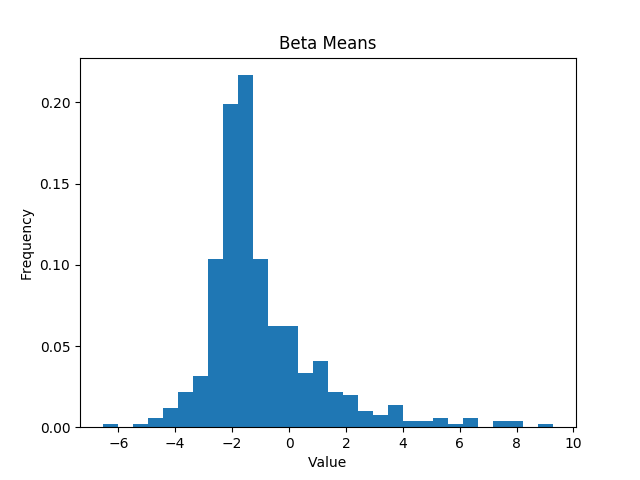}
        \end{subfigure}%
        \begin{subfigure}[b]{0.25\textwidth}
                \includegraphics[width=\linewidth]{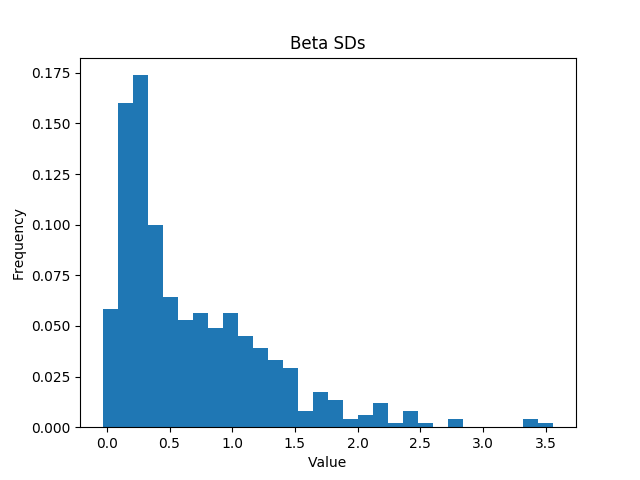}
        \end{subfigure}
        \caption{Histograms of per-channel $\gamma_{c}$ and $\beta_{c}$ statistics (mean and standard deviation) computed on CLEVR's validation set. From left to right: $\gamma_{c}$ means, $\gamma_{c}$ standard deviations, $\beta_{c}$ means, $\beta_{c}$ standard deviations. Different feature maps are modulated by FiLM in different patterns; some are often zero-ed out while other rarely are, some are consistently scaled or shifted by similar values while others by high variance values, etc.}\label{fig:film-stats}
\end{figure*}

	We visualize FiLM's effect on a single arbitrary feature map in Figures~\ref{fig:film-effect-1} and~\ref{fig:film-effect-2}. We also show histograms of per-layer $\gamma_{i,c}$ values, per-layer $\beta_{i,c}$ values, and per-channel FiLM parameter statistics in Figures~\ref{fig:gammas-layerwise},~\ref{fig:betas-layerwise}, and~\ref{fig:film-stats}, respectively.

\end{document}